\def\year{2019}\relax

\documentclass[letterpaper]{article} 
\usepackage{aaai19}  
\usepackage{times}  
\usepackage{helvet}  
\usepackage{courier}  
\usepackage{url}  
\usepackage{graphicx}  
\usepackage{navigator}
\usepackage[T1]{fontenc}    
\usepackage{booktabs}       
\usepackage{amsfonts}       
\usepackage{nicefrac}       
\usepackage{microtype}      

\usepackage{amsmath}
\usepackage{float}
\usepackage{graphicx,subfig}
\usepackage[inline]{enumitem}
\setlist[itemize]{noitemsep,topsep=-1pt,parsep=2pt,partopsep=2pt}
\usepackage[title]{appendix}
\usepackage{cprotect}
\newtheorem{example}{Example}

\def\Re{\mathbb{R}}

\def\Nat{{\rm I\kern\pIR N}}

\newcommand{\EE}[1]{\exptE\left[#1\right]}

\def\A{{\mathcal{A}}}

\def\D{{\mathcal{D}}}

\def\S{{\mathcal{S}}}

\def\vec0{{\boldsymbol{0}}}

\def\vecw{{\boldsymbol{w}}}

\def\vecz{{\boldsymbol{z}}}

\newcommand{\ra}{\rightarrow}
\newcommand{\beq}{\begin{equation}}
\newcommand{\eeq}{\end{equation}}
\newcommand{\beqa}{\begin{eqnarray}}
\newcommand{\eeqa}{\end{eqnarray}}
\newcommand{\beqan}{\begin{eqnarray*}}
\newcommand{\eeqan}{\end{eqnarray*}}
\newcommand{\ben}{\begin{eqnarray*}}
\newcommand{\een}{\end{eqnarray*}}

\def\vecw{{\boldsymbol{\bf w}}}
\def\vecz{{\boldsymbol{\bf z}}}

\renewcommand{\EE}[2]{\mathbb{E}_{#1\!\!}\left[#2\right]}

\newcommand{\CEE}[3]{\EE{#1}{{#2}~\middle\vert~{#3}}}
\renewcommand{\CEE}[3]{\EE{#1}{{#2}\mid{#3}}}

\def\CEpi#1#2{\CEE{\pi}{#1}{#2}}

\def\Epi#1{\EE{\pi}{#1}}

\def\Emj{LPj-NN}
\def\Ems{LPs-NN}
\def\TCjNN{TCj-NN}
\def\TCsNN{TCs-NN}
\def\NN{NN}
\def\TCjLin{TCj-Lin}
\def\TCsLin{TCs-Lin}

\begin{document}
%
\title{Two Geometric Input Transformation Methods for\\ Fast Online Reinforcement Learning with Neural Nets}
\author{
  Sina Ghiassian,  Huizhen Yu, Banafsheh Rafiee, Richard S. Sutton \\
  \\
  RLAI Lab, Department of Computing Science\\
  University of Alberta\\
  \texttt{\{ghiassia,huizhen,rafiee,rsutton\}@ualberta.ca}
}
\maketitle
\begin{abstract}
We apply neural nets with ReLU gates in online reinforcement learning. Our goal is to train these networks in an incremental manner, without the computationally expensive experience replay. By studying how individual neural nodes behave in online training, we recognize that the global nature of ReLU gates can cause undesirable learning interference in each node's learning behavior. We propose reducing such interferences with two efficient input transformation methods that are geometric in nature and match well the geometric property of ReLU gates. The first one is tile coding, a classic binary encoding scheme originally designed for local generalization based on the topological structure of the input space. The second one (EmECS) is a new method we introduce; it is based on geometric properties of convex sets and topological embedding of the input space into the boundary of a convex set. We discuss the behavior of the network when it operates on the transformed inputs. We also compare it experimentally with some neural nets that do not use the same input transformations, and with the classic algorithm of tile coding plus a linear function approximator, and on several online reinforcement learning tasks, we show that the neural net with tile coding or EmECS can achieve not only faster learning but also more accurate approximations. Our results strongly suggest that geometric input transformation of this type can be effective for interference reduction and takes us a step closer to fully incremental reinforcement learning with neural nets.
\end{abstract}

\section{Introduction}

Reinforcement learning systems must use function approximation in order to solve complicated real-world problems.
Neural nets provide an effective architecture for nonlinear function approximation \cite{Cyb89}, and their ability to adapt through a data-driven training process makes them powerful general function approximators for self-learning systems. Neural nets have been used since the early days of reinforcement learning \cite{MSW90,Tes95}; reinvigorated now by the advances in deep learning, they are the driving force behind most recent progresses toward large-scale reinforcement learning \cite{MnK15,Sil16,WLB18}.

On the other hand, it was also known early on that as neural nets generalize globally in training, they have a weakness: when the network gains new experience, it tends to ``forget'' what it has learned in the past, a phenomenon known as ``catastrophic interference'' \cite{McC89,Fre99}. 
Indeed, since in a neural net a function is implicitly represented by the weights on the network connections, changes in a few weights during training could result in global changes of the function. In this paper we are interested in catastrophic interference as it arises in online reinforcement learning. 


Solution methods have been proposed to address the catastrophic interference issue in the supervised learning context \cite{Fre99,Lee2017}; however, these methods are not readily available for use in the reinforcement learning framework. Most if not all of these techniques focus on multi-task supervised learning and are specific to the transfer learning context and are not applicable to the online reinforcement learning context with rapidly changing policies. The main reason behind this limitation is that interference reduction techniques proposed for supervised learning settings rely on the fact that the learning agent can successfully learn to make progress on a single task. As we will show later in the paper, due to severe interference, systems that use neural nets as function approximators can fail to make progress even on a single task in online reinforcement learning. Failing to solve a single task successfully, it is then out of the question to apply multi-task transfer-learning techniques to the reinforcement learning framework. An interesting research direction is to adapt these supervised learning techniques to the reinforcement learning context; however, this is not what we pursue in this work. Instead, we propose novel methods that are suitable for online reinforcement learning.

The main strategy to work around the catastrophic interference problem in reinforcement learning has been to train neural networks offline with batch data and experience replay buffers \cite{MnK15,Lin93}. Experience replay uses, at each training step, a rich sample of past and newly gathered data in order to update the neural network weights. This strategy seems to work in practice, but it requires a lot of memory and computation and slows down the training speed. Moreover, experience replay avoids interference at the cost of losing the advantages of online updating, which is one of the important characteristics of the reinforcement learning framework \cite{SuB18}.

In this work, we explore alternative fully incremental online approaches to mitigate the catastrophic interference problem instead of using experience replay. 
To begin with, we consider a two-layer network with one hidden layer and focus on the behavior of individual nodes that use the popular ReLU gates. We recognize that with the ReLU activation function, a neural node has to respond linearly to inputs from an entire half-space, and this global nature of ReLU gates can cause undesirable learning interference in each node's learning behavior. The observation led us to propose reducing such interferences with two input transformation methods. Both methods are geometric in nature, and as we will show, their geometric properties match well the geometric property of ReLU gates. 
As we will discuss in more detail later in the paper, both methods enable the neural nodes to respond to a local neighbourhood of their input space. This can help neural networks to generalize more locally and prevent interference. 
While input transformation is one major approach to address the interference problem in neural nets~\cite{Fre99}, the two geometric methods we study in this paper have not been considered before, to our knowledge.

The first method is tile coding~\cite{Alb75}, a classic binary encoding scheme that captures the topological structure of the input space in the codes and can help promote local generalization. We refer to the combination of tile coding with neural nets as TC-NN. We will show that compared to neural nets operating on raw inputs, TC-NN generalizes more locally, has less interference, and learns much faster. We will also show that TC-NN has advantages over the classic approach of combining tile coding with a linear function approximator (TC-Lin), especially for high-dimensional problems, in terms of function approximation capability.

The second method (EmECS) is a new method we introduce. It is based on topological embedding of the input space and geometric properties of convex sets. The idea is to embed the input space in the set of extreme points of a closed convex set, so that although with ReLU, a neural node must always respond linearly to all points from an entire half-space of the transformed input space, with respect to the original input space, it can respond only to the inputs from a small local region, thus reducing its learning interference.
As we will show, EmECS can be implemented easily and efficiently, and it differs from other high-dimensional representations in that (i) it does not increase the dimensionality of the inputs by much (indeed it can work with just one extra dimension), and (ii) it can be applied on top of any pre-extracted features that are suitable for a given task. As we will also show, EmECS shares some similarities with coarse coding \cite{HMR86}, of which tile coding is a special case, despite their being seemingly unrelated. Our experimental results show that with EmECS, neural nets can perform as well as TC-NN, achieving both fast learning and accurate approximations.

The rest of this paper is organized as follows. We first provide the background on nonlinear TD($\lambda$) and Sarsa($\lambda$). We then discuss TC-NN and EmECS methods. We present experimental results before ending the paper with a discussion on future work. A few supporting results and detailed discussions are collected in the appendices.

\section[Background]{Background: Nonlinear TD($\lambda$) and nonlinear Sarsa($\lambda$)}

\label{sct:Background}

In this paper we use TD($\lambda$) and Sarsa($\lambda$) methods~\cite{Sut88,RuN94} for solving prediction and control problems respectively. The prediction problem is that of learning the value function of a given stationary policy in a standard Markov Decision Process (MDP) with discounted or total reward criteria~\cite{Put94}. Specifically, an agent interacts with an environment at discrete time steps $t=0, 1, 2, \ldots$. If at time $t$ the agent is in state $S_t$ and selects an action $A_t$, the environment emits a reward $R_{t+1}$ and takes the agent to the next state $S_{t+1}$ according to certain probabilities that depend only on the values of $(S_t, A_t)$. We consider problems where the action space $\A$ is finite, and the state space $\S$ is either finite or a bounded subset in a Euclidean space---the problems in our experiments have continuous state spaces. A stationary policy is represented by a function $\pi:\A\times\S\ra[0, 1]$, which specifies the probabilities of taking each action at a state in $\S$. The value function of the policy $\pi$ is defined by the expected sum of the discounted future rewards (or simply the expected return), $v_{\pi}(s) = \CEpi{\sum_{k=0}^{\infty} {\gamma^k R_{t+k+1}} }{S_t=s}$, for all $s \in \S$, where $\gamma \in [0,1]$ is the discount factor and $\Epi{\cdot}$ denotes taking expectation under policy $\pi$. The prediction problem for the agent is to estimate $v_{\pi}$. In the control problem, policies are not fixed and through interactions with the environment, the agent needs to find an optimal policy that maximizes the expected return.

For prediction problems, we apply TD($\lambda$) with nonlinear function approximation to update the weights $\vecw$ of a neural network using a small step size $\alpha$ according to
\begin{equation} 
\vecw_{t+1} = \vecw_t + \alpha \, \delta_t \, \vecz_t.
\end{equation}
Here $\delta_t$ is the temporal-difference error and $\vecz_t$ the eligibility trace vector, calculated iteratively as 
\begin{align}
\delta_t &= R_{t+1} + \gamma \hat{v}(S_{t+1}, \vecw_t) - \hat{v}(S_t, \vecw_t),\\ 
\vecz_t &= \gamma \lambda \vecz_{t-1} + \nabla_{\vecw}\hat{v}(S_t, \vecw_t),
\end{align}
where $\hat{v}(s, \vecw)$ represents the approximate value for state $s$ produced by the neural net with weights $\vecw$, and $\nabla_{\vecw}\hat{v}(S_t, \vecw_t)$ denotes the gradient of the function $\hat{v}(S_t, \cdot)$ at $\vecw = \vecw_t$.

Sarsa($\lambda$) is the control variant of TD($\lambda$). When it was first proposed, it actually used neural networks as function approximators~\cite{RuN94}. Its update rules are similar to those of TD($\lambda$) except that $\hat{v}(S_t, \vecw_t)$ is replaced by $\hat{q}(S_t, A_t, \vecw_t)$, the approximate value for the state-action pair $(S_t, A_t)$ produced by the neural net. The action at each time step is typically chosen in an $\epsilon$-greedy way with respect to the current approximating function. (We use $\epsilon=0.1$.)

The network structures we used for prediction and control are different. For prediction, the network receives, as input, the state (or a representation of it) and outputs the approximate value for that state. For control, the input stays the same, but the network outputs multiple values, one for each action, to approximate the state-action values at that state. All neural nets in this work have a single hidden layer that uses ReLU gates and a linear output layer that does not use any gate function.

\section{Tile coding plus neural networks: TC-NN}
\label{sct:Tile_coding_plus_neural_networks}

\begin{figure}[t]
      \centering
      \includegraphics[width=1\linewidth]{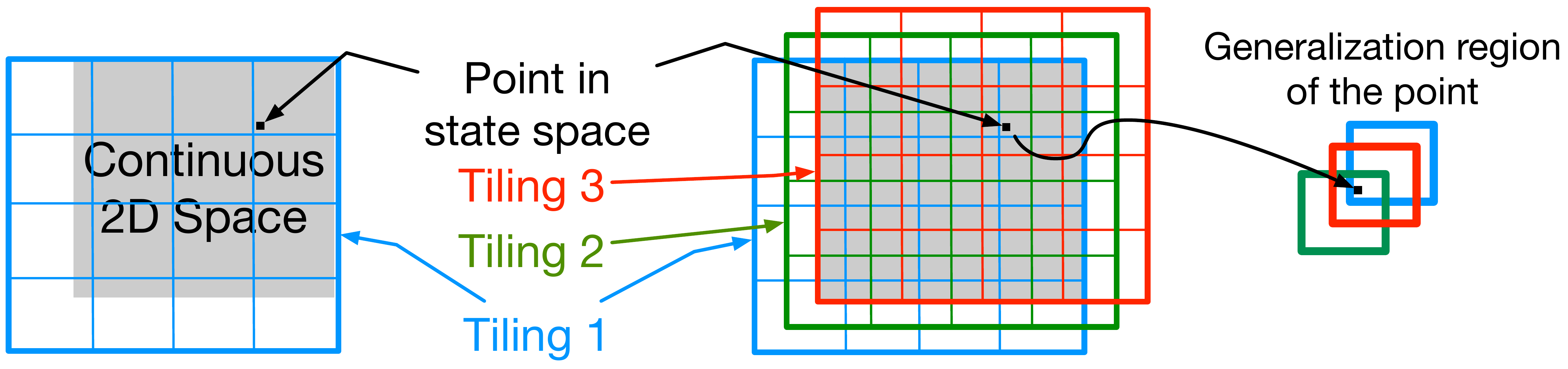}
      \caption{A continuous 2D space with 1 tiling on top of it is shown on the left. Three overlapping tilings on the 2D continuous space are shown in the middle in blue, green and red. The generalization region for a sample point is shown on the right.
      }
      \label{fig:tile_coding}
\end{figure}

Tile coding is a form of coarse coding, in which we cover the state space $\S$ with overlapping sets, and encode a state $s$ by a binary string, where the bits that are $1$ indicate which sets contain $s$. These overlapping sets capture, in a coarse way, the topological structure of the state space (i.e., which points are close together and which regions are connected to each other), and the encoding carries this structural information. In tile coding the overlapping sets are hyper-rectangles; Figure~\ref{fig:tile_coding} illustrates a simple encoding for a $2$D space. The states are thus mapped to vertices of a unit cube in a higher dimensional space. Tile coding is well-suited when the physical states occupy only a small portion of the input space, and also when the state space is non-Euclidean and has a natural product structure, as in many robotics applications. For example, in \verb+Acrobot+, two angular control parameters lie on a Cartesian product of two circles (a torus) and can be tile-coded efficiently.

Tile coding was invented by Albus \shortcite{Alb75,Alb81}. It is the key component of his CMAC computation architecture, which is, in fact,  tile coding plus a linear function approximator (TC-Lin). The nonlinear input map provided by the encoding was to facilitate \emph{local generalization}: the result of training at a particular state generalizes locally to the ``neighborhood'' of that state as defined by the union of those sets that contain the state (cf.\ Figure~\ref{fig:tile_coding}). CMAC has been applied in control and robotics and is known, among brain-inspired computational models, as a different type of neural network, an alternative to the globally generalizing, backpropagation-type neural net \cite{MGK90,BaW96}.
In reinforcement learning, Lin and Kim \shortcite{LiK91} proposed CMAC/TC-Lin for TD($\lambda$). Tham \shortcite{Tha94} used it with a variety of online reinforcement learning algorithms, including Q-learning and Sarsa($\lambda$), to solve complex robotics problems. Other successful examples of using TC-Lin with Sarsa($\lambda$) were also shown by Sutton \shortcite{Sut96}. (See the textbook \cite[Chapter 9.5]{SuB18} for an excellent introduction to tile coding and its applications in reinforcement learning.)
Given the rich history of TC-Lin, our proposal to combine tile coding with a neural net may seem unorthodox at first sight. Let us now explain the merits of this TC-NN combination, as well as its differences from TC-Lin, from several perspectives.

\begin{figure}
      \centering
      \includegraphics[width=0.9\linewidth]{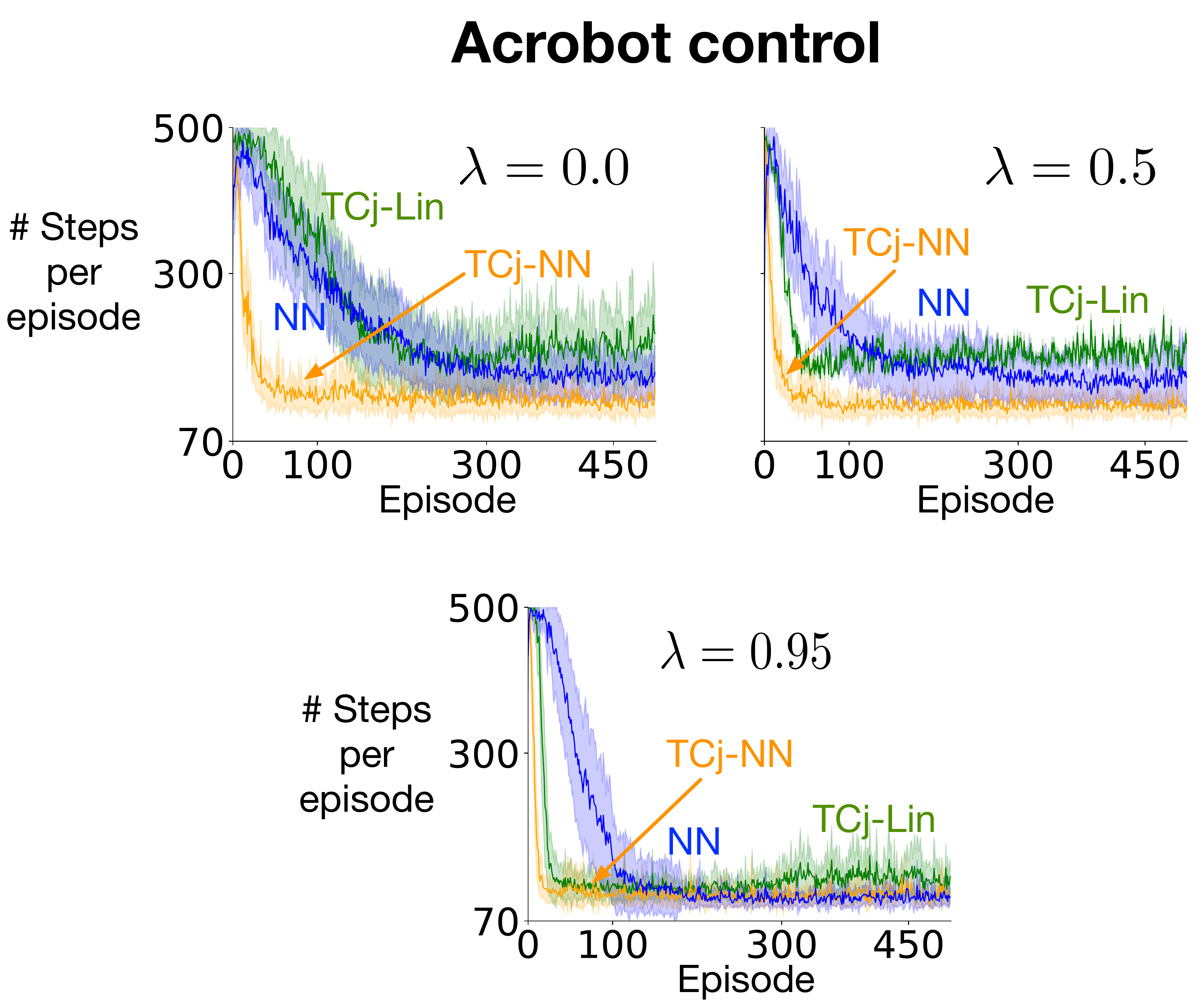}
      \caption{Learning curves for TC-Lin and TC-NN with the joint tile coding scheme (TCj-Lin and TCj-NN), and neural nets with raw inputs (NN). Standard deviation and mean of $30$ runs are shown for each curve. TCj-NN was fast and converged to a lower final performance when $\lambda=0$. As $\lambda$ got larger, different methods performed more similarly.}
      \label{fig:acrobot_tile_coding}
\end{figure}

It is true that a neural net tends to generalize globally, so in TC-NN each neural node tends to respond to a much larger area of the state space than an ideal local neighborhood as in TC-Lin. However, tile coding gives each node the ability to pick the size and shape of its activation region with respect to the original state space (see Appendix \ref{app:tile_coding} for examples of activation regions of TC-NN from an experiment). In contrast, if the neural net works on the state space directly, every node has to respond to an entire half-space linearly. This causes interference and can slow down learning considerably, as we observed in the \verb+Acrobot+ problem (Figure~\ref{fig:acrobot_tile_coding}). Sometimes the interference can be so severe that it prevents the network from learning at all (see Appendix \ref{app:nn-int} for some failure examples of neural nets with raw state inputs and with RBF features).
 
An advantage that TC-NN has over TC-Lin is in the function approximation capability. This becomes critical, as the dimensionality of the state space $\S$ increases.
To cope with the curse of dimensionality, when $\S$ has a natural Cartesian product structure, one can tile-code separately each component in the product. This encoding captures the same information as tile-coding all the dimensions of $\S$ jointly, but is much more efficient, since the resulting code length then scales linearly with the dimensionality of $\S$. However, with a linear function approximator, the encoding is also tied with how TC-Lin generalizes during training and what functions TC-Lin can approximate. As the result of these strong ties, if we tile-code each dimension separately: (i) the generalization of TC-Lin becomes global, and (ii) the set of functions TC-Lin can approximate becomes limited, since it can represent only functions that are sums of functions of each component.
In contrast, for TC-NN, if we use the separate tile coding scheme: (i) the neural net still has the freedom to choose regions of generalization as before, and these regions need not be as global as those in TC-Lin, and (ii) the set of functions that the neural net can approximate remains the same. The latter is because with either the separate or joint tile coding scheme, the states are mapped to vertices of a hyper-unit-cube with the same granularity, and the neural net can separate each vertex from the rest of the vertices by using a single hidden node (with ReLU) and assign a specific value to that vertex. We will show experimental results that confirm this advantage of TC-NN in the experimental results section (cf.\ Figure~\ref{fig:MCC_MCP_joint_vs_separate}) and in Appendix \ref{app:tile_coding} (cf.\ Figure~\ref{fig:joint_vs_separate_all_lambda}), where we will also discuss this subject in a more intuitive manner.

\section{Embedding into Extreme points of a Convex Set (EmECS)}
\label{sct:EmECSMethod}

\begin{figure}[t]
  \centering
   \includegraphics[width=0.45\textwidth]{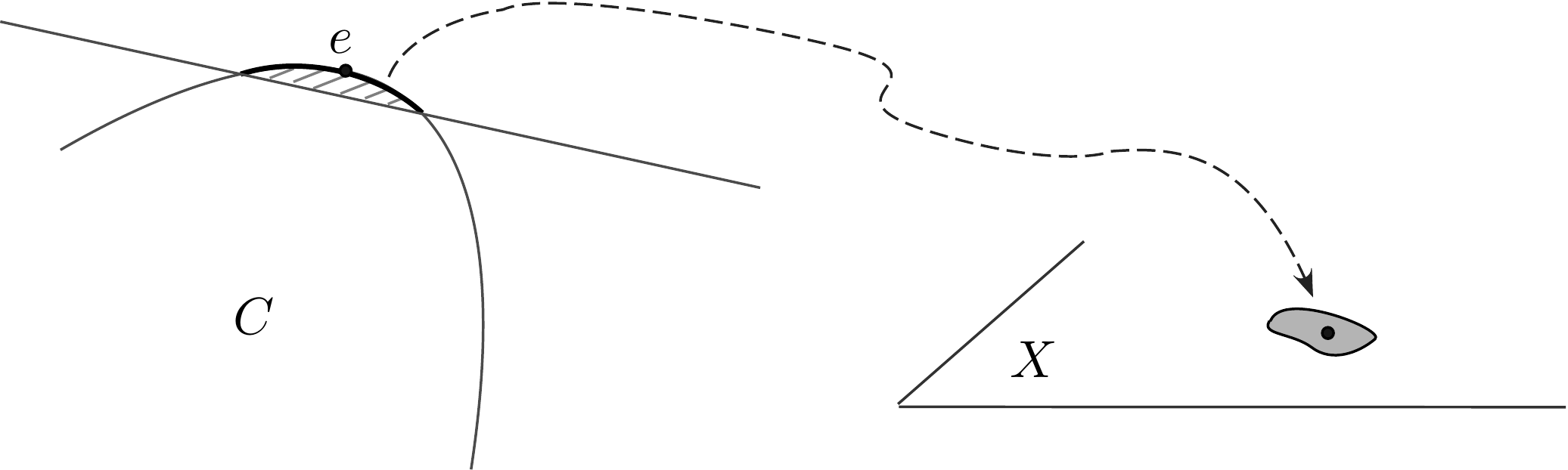} 
   \caption{On the left is a cross-section view of a hyperplane cutting out a small part of a closed convex set $C$ (in $\Re^3$) around an extreme point $e$ (black dot). This point $e$ corresponds to the point $f^{-1}(e)$ in the original input space $X$ shown on the right. The entire half-space above the hyperplane is the activation region of a neural node in the transformed input space. But only those boundary points of $C$ above the hyperplane (indicated by the thick black line in this cross-section view) correspond to real inputs in $X$, which form a small neighborhood (shaded area) around the point $f^{-1}(e)$ (black dot). The neural node thus only responds to inputs from that neighborhood in $X$.}
   \label{fig:emecs1}
\end{figure}

We now introduce a new input transformation method, EmECS, for reducing the learning interference of individual neural nodes with ReLU gates. 
This method is based on two geometric properties: 
\begin{itemize}[leftmargin=18pt]
\item[(i)] With ReLU, the activation region of a neural node is the open half-space 
\begin{equation}  \label{eq-hsp}
 \{ x \mid \langle w, x \rangle + b > 0 \}
\end{equation} 
that corresponds to the hyperplane $\langle w, x \rangle + b = 0$, 
where $w$ is the vector of weights and $b$ the scalar bias term associated with the node.
\item[(ii)] For a closed convex set $C$, consider a point $x \in C$ and the neighborhoods of $x$ relative to $C$ (i.e., the intersections of its neighborhoods with $C$). If $x$ is an extreme point of $C$, there is a hyperplane whose open half-space (\ref{eq-hsp}) contains only a (arbitrarily) small neighborhood of $x$ relative to $C$.
\footnote{An \emph{extreme point} of a convex set $C$ is one that cannot be expressed as a convex combination of other points of $C$. For a closed convex set $C$, by Straszewicz's Theorem \cite[Theorem 18.6]{Roc70}, every extreme point is the limit of some sequence of exposed points, where an \emph{exposed point} of $C$ is a point through which there is a supporting hyperplane that contains no other points of $C$ \cite[Section 18, p.\ 163]{Roc70}. This means that for an exposed point $y$, there is a linear function $f_y$ achieving its maximum over $C$ uniquely at $y$. Consequently, for any extreme point $x$, we can choose an exposed point $y$ sufficiently close to $x$ so that for some linear function $f_y$ with the property just mentioned and for some $\epsilon > 0$, the half-space $\{ z \mid f_y(z) \geq f_y(y) - \epsilon\}$ contains only a small neighborhood (relative to $C$) of $x$. As this neighborhood of $x$ consists of the $\epsilon$-optimal solutions of $\sup_{z \in C} f_y(z)$, it can be made arbitrarily small by the choices of $y$ and $\epsilon$. The corresponding linear function $f_y$ then gives the hyperplane $f_y(z) = f_y(y) - \epsilon$ with the desired property (ii), thus proving our claim.}
The left part of Figure~\ref{fig:emecs1} illustrates this property of an extreme point.
\end{itemize}

If $X$ is the original input space of the neural net ($X$ can be the state space of the problem or the space of any pre-extracted features of states), our method is to embed $X$ in the set of extreme points of a closed convex set in a higher dimensional Euclidean space, and let the neural-net work with the transformed inputs instead of the original inputs. 

Here, by embedding, we mean a one-to-one continuous map $f$ whose inverse $f^{-1}:$ $f(X) \to X$ is also continuous. Such a map is called a topological (or homeomorphic) embedding because it preserves topological properties~\cite{Eng89}. For example, if the states lie on a manifold in $X$ (say, a torus), their images under $f$ lie on a topologically equivalent manifold (thus also a torus), and if the states form two disconnected sets in $X$, so do their images under $f$. 
By combining this topology-preserving property of an embedding with 
the geometric properties of convex sets discussed earlier, we obtain the following. If we choose a closed convex set $C$ whose boundary points are all extreme points, and if we embed $X$ into the boundary of $C$, then a neural node with a ReLU gate, when applied to the transformed inputs, becomes capable of responding to only a small neighborhood of any given point in the original input space $X$ (cf.\ Figure~\ref{fig:emecs1}). This explains the mechanism of our EmECS method: it enables each neural node to work locally, despite the global nature of ReLU.

Of course, having the ability of localized generalization at each node does not mean that the network always allocates a small region to each node or different regions to different nodes---indeed it is hard for such coordination between nodes to emerge automatically during training. Nonetheless, our experiments showed that EmECS can improve considerably neural nets' learning performance.

We can implement EmECS efficiently. Below are a few simple examples of the embedding; in our experiments we have used (a) and (c) (which give the LPj-NN and LPs-NN algorithms in the experimental results section). 
\begin{example}[Some instances of maps for EmECS] \rm \label{ex-1}
Suppose $X \subset \Re^n$.
\begin{itemize}[leftmargin=18pt]
\item[(a)] Map $X$ into an $n$-sphere of radius $r$ in $\Re^{n+1}$ by first ``lifting'' the set $X$ along the $(n+1)$-th dimension and then projecting it on the sphere. Specifically, for $x \in X$, let
\begin{equation}
f(x) = r \cdot (x, 1)/\| (x, 1)\|_2.
\end{equation}
We shall refer to this type of map as lift-and-project (LP for short).
\item[(b)] Let $g$ be a continuous, strictly convex function on $\Re^n$ (e.g., $g(\cdot) = \| \cdot\|_2^2$). Map $x \in X$ to
$f(x) = (x, g(x)).$
This embeds $X$ into the graph of the function $g$, and the closed convex set $C$ here is the epigraph of $g$: $C = \{(x, \nu) \mid x \in \Re^n, \, \nu \in \Re, \, \nu \geq g(x) \}$.
\item[(c)] If $X = X_1 \times X_2 \times \cdots \times X_m$ where each $X_i \subset \Re^{n_i}$, we can separately embed each $X_i$ in $\Re^{n_i+1}$, with a map $f_i$ of the form given in (a)-(b), for instance. 
The result is the embedding of $X$ in $\Re^{n+m}$ given by
$f(x) = \big(f_1(x), \ldots, f_m(x)\big).$
The range of $f$ is a subset of extreme points of the closed convex set $C = C_1 \times \cdots \times C_m$, where $C_i$ is the convex set associated with the embedding $f_i$.
Sometimes, a component space $X_i$ already contains the desired embedding of the state components (e.g., when the latter lie on a circle or sphere in $X_i$). Then we do not need to embed $X_i$ any more and can simply take $f_i$ above to be the identity map $f_i(x_i) = x_i$.
\end{itemize}
\end{example}

EmECS shares some similarities with coarse coding~\cite{HMR86} despite their being seemingly unrelated. With EmECS, the activation regions of individual nodes, viewed in the original input space, resemble the receptive fields (i.e., the overlapping sets) in coarse coding. Like the latter, each activation region is connected (thanks to the embedding property), if the hyperplane associated with the node has its normal vector point in the right direction in the transformed input space. For instance, for the lift-and-project map in Example~\ref{ex-1}(a), it suffices that the normal vector points ``upwards'' with respect to the extra $(n+1)$-th dimension (in our experiments we always initialize the network weights in this way). In coarse coding, receptive fields can have different sizes and shapes, but they are chosen before learning takes place. With EmECS, the activation regions of neural nodes change their locations and sizes dynamically during learning. The shapes of these regions depend partly on the embedding, so by choosing the embedding, we can have some influence over them, like in coarse coding. For example, the separate embedding scheme in Example~\ref{ex-1}(c) gives the network more freedom to produce activation regions that are asymmetric, wider in some dimensions and narrower in others, whereas a joint embedding scheme like Example~\ref{ex-1}(a) can be used to enforce symmetric or specific asymmetric shapes. (See Appendix \ref{app:emex-cc} for illustrations and a more detailed discussion.) 

Let us now compare EmECS and TC-NN.
They both map the original inputs to the extreme points of a convex set---in the case of tile coding, the convex set is a hypercube and the extreme points are its vertices, and they both use the topological structure of the original space to do so. A difference between them is that for EmECS the input transformation is an embedding, whereas for tile coding it is not. As a consequence, in TC-NN, an activation region of a neural node, viewed in the original input space, can (and usually do, as observed in our experiments) contain multiple disconnected components. This suggests that one may be able to further improve the performance of TC-NN by initializing the neural net in a certain way or by monitoring and ``pruning'' the activation regions of its nodes during training.
Another difference between the two methods is in computational complexity. For TC-NN, suppose each dimension of the inputs can be tile-coded separately; then the dimensionality of the transformed inputs will still depend on the size of the original input space along each dimension. In contrast, EmECS only increases the dimensionality of the inputs by the number of component spaces that are embedded separately (cf.\ Example~\ref{ex-1}). So, given the same number of hidden-layer nodes, the neural net with EmECS has much fewer parameters than the TC-NN network.

\section{Experimental results}
\label{sct:Experimental_results}

In this section, we show experimentally that our proposed methods are fast and accurate. We compare our proposed methods with two existing online methods: tile coding plus linear function approximation (TC-Lin) and neural networks with raw inputs (\NN{}). We do not compare our methods to experience replay as it is not a fully incremental online method. Our proposed methods are tile coding plus neural networks (TC-NN) and EmECS plus neural networks. From EmECS, we used the lifting-and-projecting scheme plus neural networks (LP-NN). We add letters j and s to TC and LP (e.g., \TCjNN{} or \TCsNN{}) to show whether the dimensions of the input are transformed in a joint or in a separate fashion. 
\Emj{} uses the lift-and-project map in Example~\ref{ex-1}(a), and \Ems{} uses the separate embedding scheme in Example~\ref{ex-1}(c) with each component map $f_i$ being a lift-and-project map. 
We first use three small problems to compare \TCjNN{}, \Emj{}, \NN{}, and \TCjLin{}: \verb+Mountain Car+ prediction, \verb+Mountain Car+ control and \verb+Acrobot+ control. All problems are on-policy, undiscounted, and episodic. We perform another set of experiments on the \verb+Mountain Car+ prediction problem to study the effects of transforming the dimensions of the input jointly or separately. Finally, we assess the practicality of our methods in higher dimensions by applying \TCsNN{} and \Ems{} to a real world robot problem in an off-policy continuing setting. We also present (in the appendix on learning interferences in online training of neural nets) the results of using RBF kernels to transform the input space and show that not every input transformation method that uses the neighborhood information or creates sparse features can be effective. Implementation details of these experiments are given in Appendix \ref{app:Experimental_details}.

\begin{figure}[th]
      \centering
      \includegraphics[width=1\linewidth]{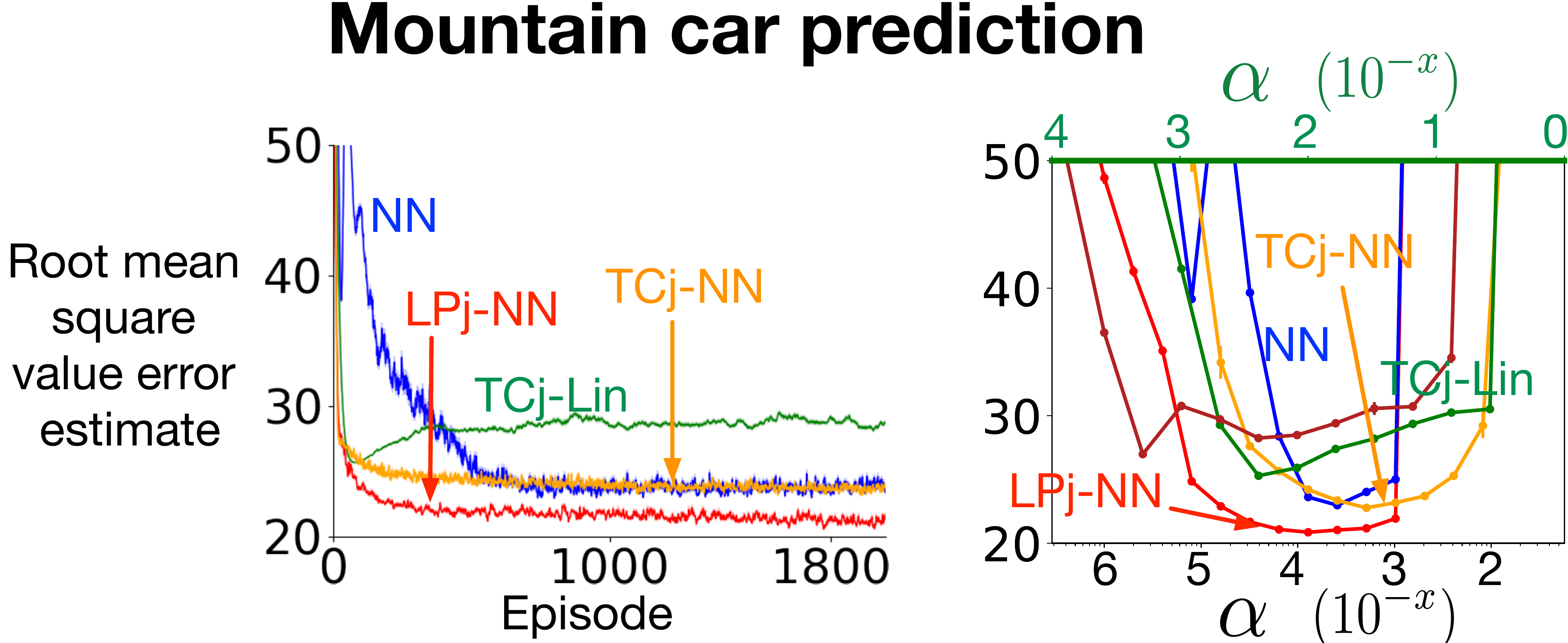}
      \caption{Learning curve (left) and parameter study (right) using TD(0). The top and bottom axes in the parameter study show the step size values for linear function approximation and neural nets respectively. TCj-Lin converged to a higher asymptotic performance compared to other methods. Neural networks alone (without any input transformation) slowly converged to a good final performance. The proposed methods were fast and converged to a low final performance.}
      \label{fig:MCP}
\end{figure}

\begin{figure*}[th]
      \centering
      \includegraphics[width=0.8\linewidth]{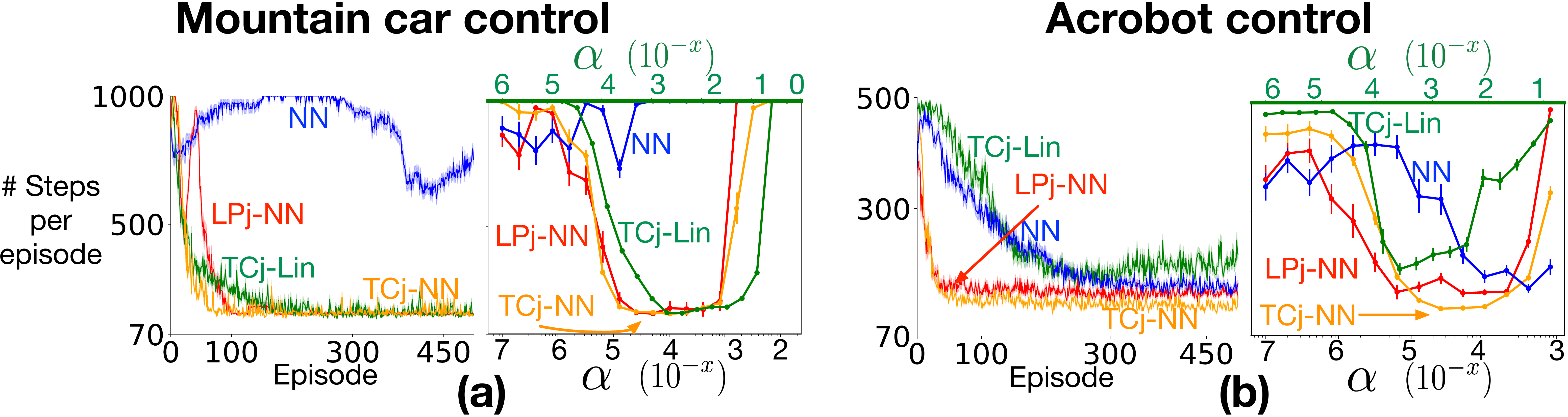}
      \caption{Results on two control problems with $\lambda=0$. Proposed methods were fast and converged to a good asymptotic performance on both problems. Neural nets alone (without any input transformation) were unable to solve the mountain car problem due to catastrophic interference.} 
      \label{fig:acrobot_MCControl_mains}
\end{figure*}
The first testbed was \verb+Mountain Car+ in the prediction setting, involving the evaluation of a fixed policy. 
The policy was to push towards the direction of velocity. Neural nets that were used with different input transformation methods, had different numbers of inputs and hidden units. \NN{} had 2 raw inputs: position and velocity. \Emj{} had 3 inputs: position, velocity, and its extra dimension. \TCjNN{} used a feature size of 80 (see Appendix \ref{app:Experimental_details} for details of tile coding and why the number of features is 80). \TCjLin{} had the same number of features as \TCjNN{}. In this problem we carefully chose the number of hidden units to make sure all methods had almost the same number of weights in the neural net. We gave \NN{} and \Emj{} 135 and 100 hidden units to create networks with a total of 405 and 400 weights respectively. We gave \TCjNN{} only 5 hidden units, which resulted in a network with 405 weights.

We ran each method under each parameter setting for 30 independent times (30 runs). Each run had 2000 episodes. We then averaged over runs to create learning curves. We also performed a parameter study over different step sizes: for each $\alpha$ and each run, we computed an average over the last 5\% of episodes, which produced 30 numbers -- one for each run. We then computed the mean and standard error over the resulting numbers. We used our parameter study results to choose the value of the step size for the learning curves we plotted. For all methods, we chose the largest step size (and thus fastest convergence) for which the final performance was close to the best final performance of that method. We used an estimation of the root mean square value error as the error measure:
\[\widehat{\text{RMSVE}}(\vecw_t)  =\sqrt{ \frac{1}{|\mathcal{D}|} \sum_{s \in \mathcal{D}} \left[\hat{v}(s, \vecw_t) - v_{\pi}(s)\right]^2}\]
Here $\mathcal{D}$ is a set of states that is formed by following $\pi$ to termination and restarting the episode and following $\pi$ again. This was done for 10,000,000 steps, and we then sampled 500 states form the 10,000,000 states randomly. The true value $v_\pi(s)$ was simply calculated for each $s \in \D$ by following $\pi$ once to the end of the episode.

Results on the \verb+Mountain Car+ prediction (Figure \ref{fig:MCP}) show that \NN{} had a good final approximation of the value function; however, it was slow. \TCjLin{} was fast but it could not approximate the value function as accurately as other methods. \TCjNN{} and \Emj{} were both fast and approximated the value function accurately. \Emj{} made the most accurate approximation.

The second testbed was \verb+Mountain Car+ in the control setting. We used $\epsilon$-greedy Sarsa($\lambda$) with $\epsilon=0.1$. The performance measure was the number of steps per episode, which is equal to the negative return. We did 30 runs. Each run had 500 episodes. \NN{}, \TCjNN{} and \Emj{} all had 800 hidden units. We used the same tile coding scheme as in the \verb+Mountain Car+ prediction problem.

All methods except \NN{} learned to successfully reach the goal. \NN{} could not solve the task with raw inputs. \TCjNN{} was the fastest method to achieve its best final performance, \Emj{} came second and \TCjLin{} was the slowest (see Figure \ref{fig:acrobot_MCControl_mains}(a)).

The third testbed was \verb+Acrobot+ in the control setting. We used $\epsilon$-greedy Sarsa($\lambda$) with $\epsilon=0.1$. Performance measure was similar to \verb+Mountain Car+ control. \TCjLin{} and \TCjNN{} had a feature size of 256 (see Appendix \ref{app:Experimental_details} for tile coding details). \TCjNN{} used 4000 hidden units. \NN{} and \Emj{} had 2000 hidden units. \Emj{} fed the neural net with 5 inputs (original dimensions plus one extra dimension). We did 30 runs. Each run had 500 episodes. 

The best final performance was achieved by \TCjNN{}, followed by \Emj{}, then \NN{} and then \TCjLin{}. Speed wise, \Emj{} and \TCjNN{} were the fastest methods. Figure \ref{fig:acrobot_MCControl_mains}(b) summarizes the results.

\begin{figure}[h]
      \centering
      \includegraphics[width=0.9\linewidth]{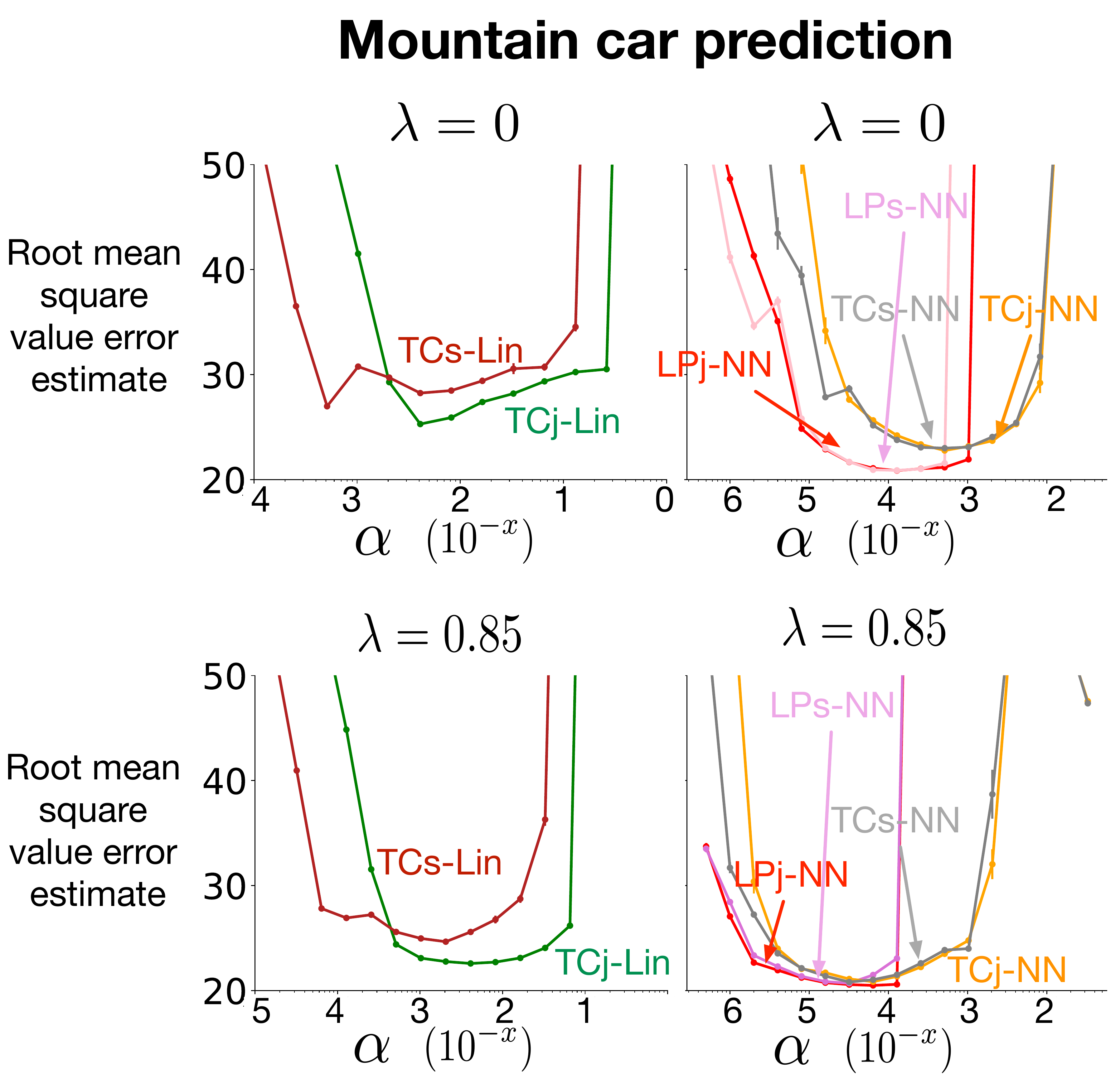}
      \caption{Parameter studies comparing separate and joint input transformation. Transforming the input dimensions jointly or separately did not affect the performance of the proposed methods; however, it affected the performance of tile coding plus linear function approximation.}
      \label{fig:MCC_MCP_joint_vs_separate}
\end{figure}

We also did a parameter study on the \verb+Mountain Car+ prediction problem to study the effects of transforming the dimensions of the input jointly and separately. The final performances of \Ems{} and \Emj{} were similar, so were the final performances of \TCjNN{} and \TCsNN{}. However, the final performance of \TCsLin{} was worse than that of \TCjLin{}. This confirms one of our assumptions from Section~\ref{sct:Tile_coding_plus_neural_networks}: tile coding the input dimensions separately (vs jointly) does not pose generalization restrictions (and does not affect the final performance) when combined with neural nets. However, it does pose restrictions (and affects the final performance) if combined with linear function approximation. See Figure~\ref{fig:MCC_MCP_joint_vs_separate} for these results when $\lambda = 0$ and $0.85$, and see Appendix \ref{app:tile_coding} for results on other values of $\lambda$.

As a starting point for working in higher dimensions, we applied our methods to a real world robot task in the off-policy setting. In this problem, a Kobuki robot wanders in a pen, learning to predict when it will bump into something if it goes forward. More details about the reward and policies can be found in Appendix \ref{app:Experimental_details}. The sensory information available to the robot to learn this task consisted of 50 RGB pixels from its camera, represented as a vector of size 150. We did 30 runs of 12000 time steps and used an estimation of root mean square return error as the performance measure:

\[\widehat{\text{RMSRE}}(\vecw_t)  =\sqrt{ \frac{1}{|\mathcal{D}|} \sum_{(s, G) \in \mathcal{D}} \left[\hat{v}(s, \vecw_t) - G\right]^2}\]
Here $\mathcal{D}$ is a set of state and return pairs selected according to the following procedure. To sample each pair $(s, G)$, the robot followed the behavior policy for some random number of steps, sampled state $s$, and followed the target policy from $s$ to compute the true return $G$. After sampling each pair, the robot switched back to the behavior policy for a random number of time steps to get the next sample. We repeated this whole procedure 150 times to construct $D$.

\begin{figure}[th]
      \centering
      \includegraphics[width=1\linewidth]{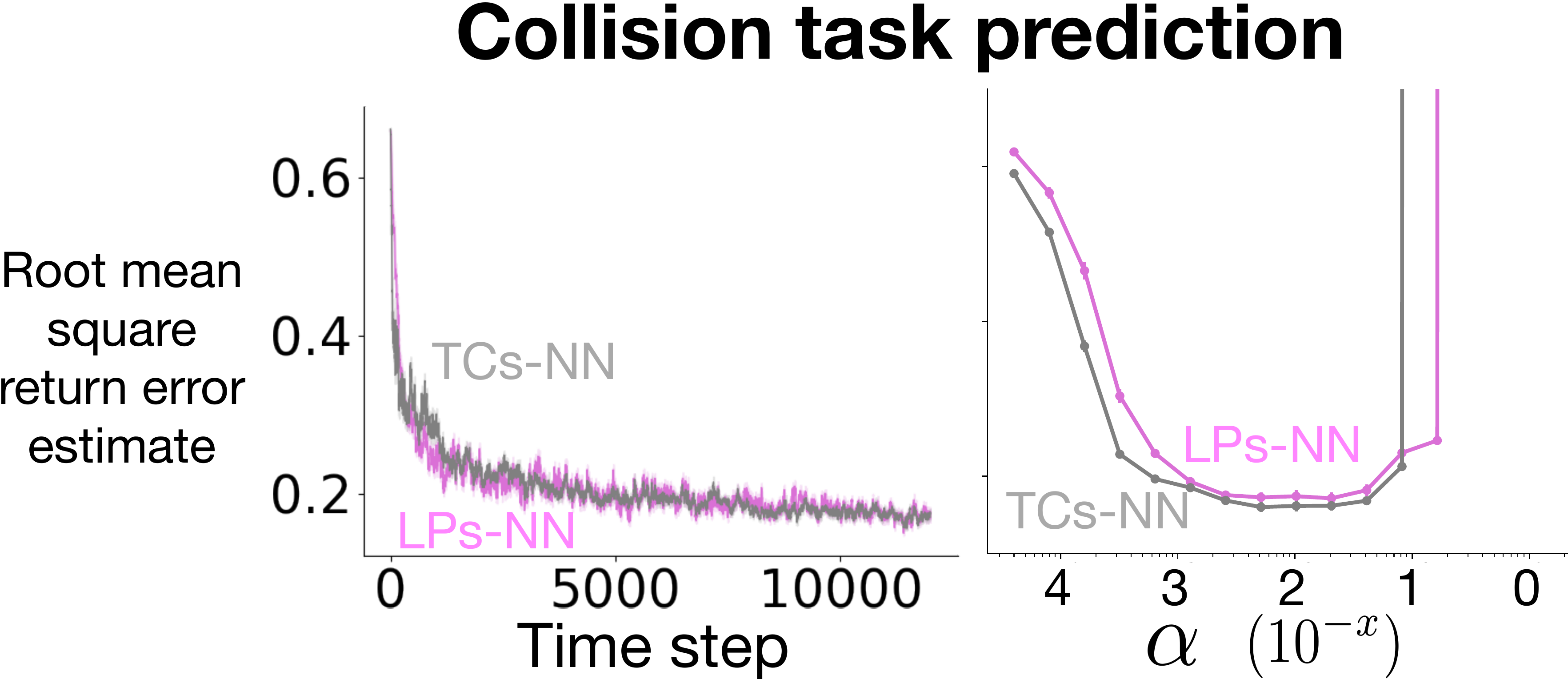}
      \caption{Learning curve and parameter study for our proposed methods on the robot collision task. Results show that our methods can be effective in higher dimensional spaces.}
      \label{fig:Collision}
\end{figure}

We tile coded each of 150 numbers separately. Tile coding produced a feature vector of size 9600. \Ems{} had 300 features. Both \TCsNN{} and \Ems{} had 1000 hidden units. More experimental details can be found in Appendix \ref{app:Experimental_details}. Our methods worked well in this environment. The results are presented in Figure \ref{fig:Collision} (\TCsLin{} and \NN{} also performed well; their results are not shown here).

We also studied the effect of using larger values of $\lambda$. Eligibility traces has been shown to be effective in the past \cite{RuN94,Sut96}. Our results (in Appendix \ref{app:largerLambda}) confirm that larger values of $\lambda$ help all methods (except RBFs) to learn faster and more accurately. One of the reasons can be that eligibility traces carry the past information and this can help prevent interference. Eligibility traces are not computationally expensive and can be an alternative to experience replay.

\section{Conclusions and future work}
\label{sct:Discussion}

In this paper, we took a step towards understanding the catastrophic interference problem as it arises in online reinforcement learning. We showed that the two geometric input transformation methods, tile coding and EmECS, can help improve the online learning performance of single hidden layer neural nets with ReLU gates. Reinforcement learning systems using these transformation techniques can learn, in a fully incremental online manner, to successfully accomplish a task. While our focus was on reinforcement learning, these two methods can also be applied in supervised learning with neural nets. 

Future work is to develop both methods further and understand them better.
Our ongoing research includes: to experiment with different embeddings for EmECS and study their effects; to test both methods on larger problems; and to use them with multilayer neural nets and hierarchical neural nets, for reducing the learning interference of individual nodes at higher hidden layers of these networks. A simple idea for combining EmECS and neural nets with multiple hidden layers is to apply EmECS to each hidden layer. Experimenting these methods with other activation functions such as other variants of ReLU and sigmoidal functions can also lead to interesting observations.

Another interesting future research direction is to try to understand if proposed methods can make improvements on a system that uses experience replay. Since experience replay uses batch data, it is unfair to compare it directly with online reinforcement learning algorithms. However, a more meaningful comparison could be adding experience replay to the current methods and comparing the resulting methods with a system that uses experience replay to overcome interference.

Our proposed methods provide a new point of view on the problem of interference and its possible solutions, and there is a wide range of research directions that can be pursued.
\subsubsection*{Acknowledgments} 
The authors gratefully acknowledge funding from Alberta Innovates--Technology Futures, the Natural Sciences and Engineering Research Council of Canada, and Google DeepMind.
\clearpage

{
\small
\bibliographystyle{aaai}
\bibliography{tc_emex_bib}
}

{\clearpage
\addcontentsline{toc}{section}{Appendices}
\appendix
\appendixpage
\renewcommand{\setthesection}{\appendixname \Alph{section}
}

\section{Experimental details}
\label{app:Experimental_details}

For implementing neural networks in different settings we used the PyTorch software \cite{Pa17PyTor}.

\paragraph{Mountain Car prediction:}The original \verb+Mountain Car+ problem has a 2-dimensional state space: position and velocity. The position can vary between $-1.2$ and $0.6$, and the velocity varies between $-0.07$ and $0.07$. There are three actions in each state, full throttle forward, full throttle backward and no throttle. The car starts around the bottom of the hill randomly (uniformly chosen position between $-0.4$ and $-0.6$). The reward is $-1$ for each time step before getting to the goal state at the top of the hill when the position becomes larger than $0.5$.

Our fixed-policy \verb+Mountain Car+ testbed used the same environment as the original one only with a fixed policy. The policy was to push towards the direction of velocity.

We applied four methods along with nonlinear TD($\lambda$) to approximate the value function of the policy. Each method had different parameters. \TCjLin{} and \TCjNN{} both used 5 tilings, each of which had $4 \times 4$ tiles. \NN{} used normalized (between $-1$ and $1$) raw inputs as its input to the neural net. 

\Emj{} normalized the input space (between $-1$ and $1$) and then lift-and-projected the input space (as explained in Example \ref{ex-1}(a)) to generate transformed features for the neural network. For this method we used a radius $r=8$ in the lift-and-project map. After performing the lift-and-project, we shifted the origin in the extra dimension by $6$ so that the hyperplanes associated with hidden units of the neural network are close to the transformed inputs. This made it easier for each node to respond to a specific part of the transformed space. For \Emj{} and \Ems{} we made sure that the normal vectors of the hyperplanes created by neural net weights are initialized such that they pointed ``upwards'' with respect to the extra dimensions (this is done for all experiments).

The network's weights for all methods were initialized using a normal distribution with mean 0 and standard deviation $0.5$. We initialized the biases by drawing random numbers from a normal distribution with mean $0$ and standard deviation $0.1$.

\paragraph{RBF and sparse RBF features:}We also used RBF kernels in the \verb+Mountain Car+ prediction problem to transform the input features. RBF features are created using the following formula:
\[x_i(s)=\text{exp}\left( - \frac{||s-c_i||^2}{2\sigma_i^2} \right) \]
where $x_i$ is the $i$th RBF feature, $c_i \in \S$ is the center and $\sigma_i>0$ is the width. In our experiments we used 50 and 100 centers. These centers were chosen randomly uniformly in the state space. We used two different widths (with respect to normalized input): $0.085$ and $0.1$. (The results for $0.085$ are not presented in paper because they were always worse that the results for $0.1$.) The neural nets were initialized as before. For each state in the state space, its RBF features were first computed with respect to all of the centers (which creates a feature of size $50$ or $100$ in our case) and then used as the input to the neural net. 

We also used another version of RBF which we call sparsified RBF. In this version, we found the features that are smaller than $0.001$ and set them to $0$. This process made the input sparse. We refer to this method in figures as SRBF.

\paragraph{Mountain Car control:}We again used the classic \verb+Mountain Car+ domain but this time in the control setting. We used the same number of tiles and tilings in this experiment. \Emj{} also used the same parameters as in the prediction case. Weights and biases of the networks for all methods were initialized with a normal distribution with mean $0$ and standard deviation $0.1$. In this problem, if the episode took more than 1000 steps, it was terminated manually.

\paragraph{Acrobot:}The \verb+Acrobot+ is similar to a gymnast. Its goal is to swing its feet above the bar its hanging from, in order to end the episode successfully. A state consists of two angles and two angular velocities. There are three actions: positive, negative, and no torque. The reward is $-1$ for each time step before the episode ends.

We used the \verb+Acrobot+ problem from Open AI Gym \cite{Brock16}. We created the tile coded features for this problem using the freely available tile coding software from Richard Sutton's web page. We used $8$ tilings, each of which was of size $2^4$ since \verb+Acrobot+ has $4$ dimensions, $2$ tiles for each (we used a memory size of $256$). All neural networks were initialized the same way as in the \verb+Mountain Car+ control problem. \Emj{} used the same parameters as in the previous problems. If the episode took more than $500$ steps, it was terminated manually.

\paragraph{Robot prediction problem:} Here we provide the details on the robot prediction problem (which we also refer to as the collision task). The robot wants to learn when it will bump into something if it goes forward. If the robot's sensors detected a bump, the reward became $1$ and $\gamma$ became $0$, otherwise reward and $\gamma$ took the values of $0$ and $0.97$ respectively. The target policy was to go forward in all states. The behavior policy was to go forward 90\% of the time and turn left 10\% of the time.

We used the same tile coding software that we used for \verb+Acrobot+. To tile code each of the pixels, we used $8$ tilings, each of which has $4$ tiles (and a memory size $64$). The size of the final feature vector was $9600$. 

\Ems{} used a radius $r=3$ in the lift-and-project map. After lifting-and-projecting the input, we shifted the origin in every extra dimension by $2$. We initialized all weights and biases for all neural networks of different methods with a normal distribution of mean $0$ and standard deviation $0.01$.

\clearpage
\section{Learning interferences in online training of neural nets} \label{app:nn-int}

On the \verb+Mountain Car+ problem, neural nets operating on the raw state inputs suffered severe learning interferences during online training. In most cases, they failed to learn at all; occasionally, they succeeded but only after a long period of training. A sample of six learning curves is plotted in Figure~\ref{fig:nn_learning_int1} to illustrate this behavior. Each curve corresponds to one run of Sarsa($\lambda$) with $\lambda=0.85$. The network parameters are the same for all six curves; in particular, the hidden layer has $800$ units, and the step size $\alpha=10^{-5}$.

\begin{figure}[h]
  \centering
   \includegraphics[width=0.78\linewidth]{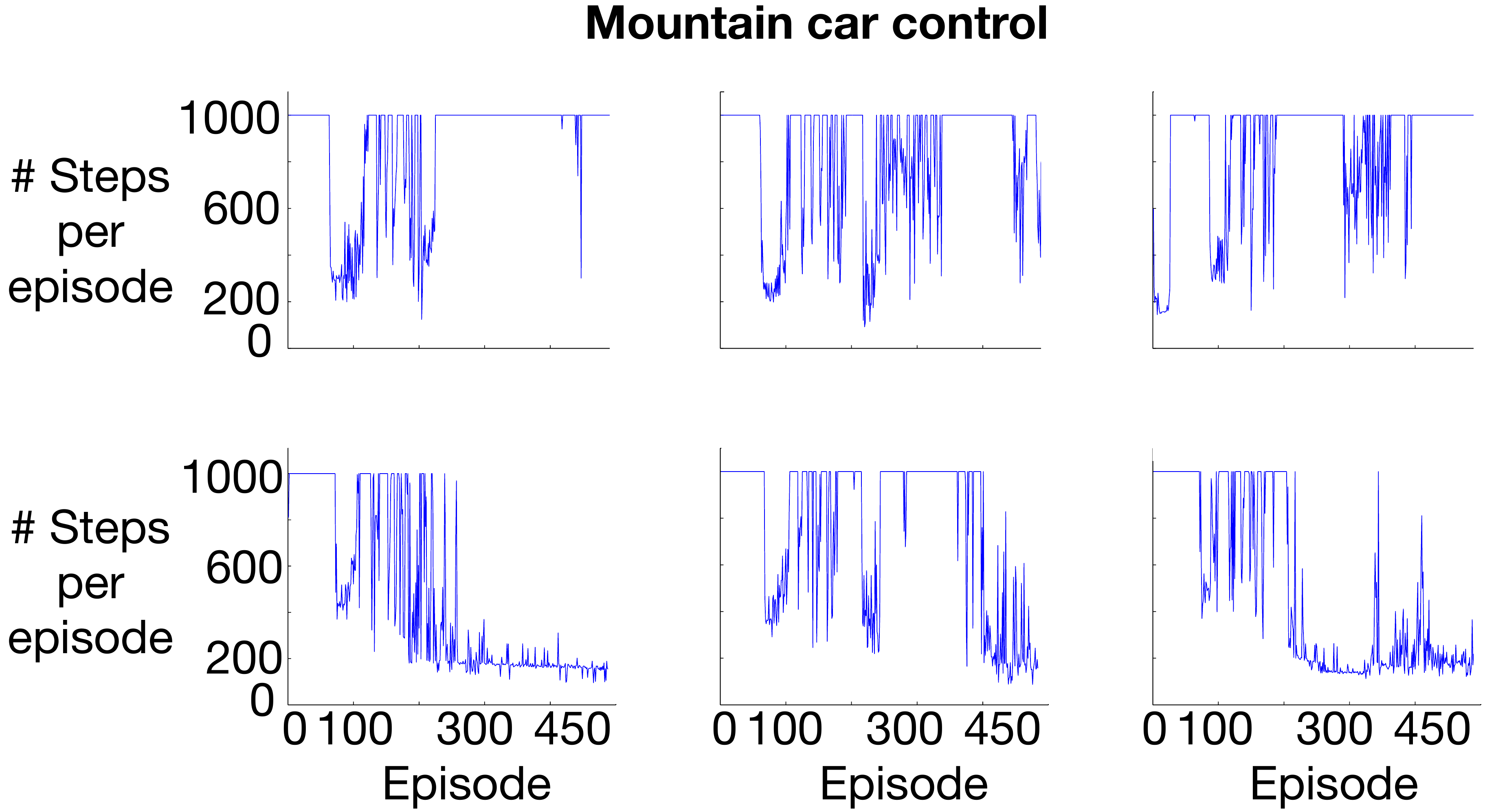} 
   \cprotect\caption{Learning interferences occurred in online training of neural nets with raw state inputs, on the \verb+Mountain Car+ problem. Top row: three failed runs; bottom row: three partly successful runs.}
   \label{fig:nn_learning_int1}
\end{figure}   

We also observed learning interferences in neural nets that operate on RBF features, on the \verb+Mountain Car+ control problem.
Figure~\ref{fig:nnrbf_learning_int} plots the learning curves of six sample runs of Sarsa($0$), 
where the neural net has $800$ hidden units and operates on $100$ RBF features that are generated with width parameter $0.1$ (with respect to the normalized inputs), and the step size used is about $0.0002$, one of the best step sizes based on our parameter study. A similar sample of six learning curves for neural nets with $100$ sparsified RBF features is plotted in Figure~\ref{fig:nnsrbf_learning_int}, where the step size used to obtain these curves is about $2.6 \times 10^{-5}$, one of the best choices according to our parameter study, and sparsified RBF features are created from RBF features by replacing small values by $0$ (see Appendix~\ref{app:Experimental_details} for details).

\begin{figure}[h]
  \centering
   \includegraphics[width=0.78\linewidth]{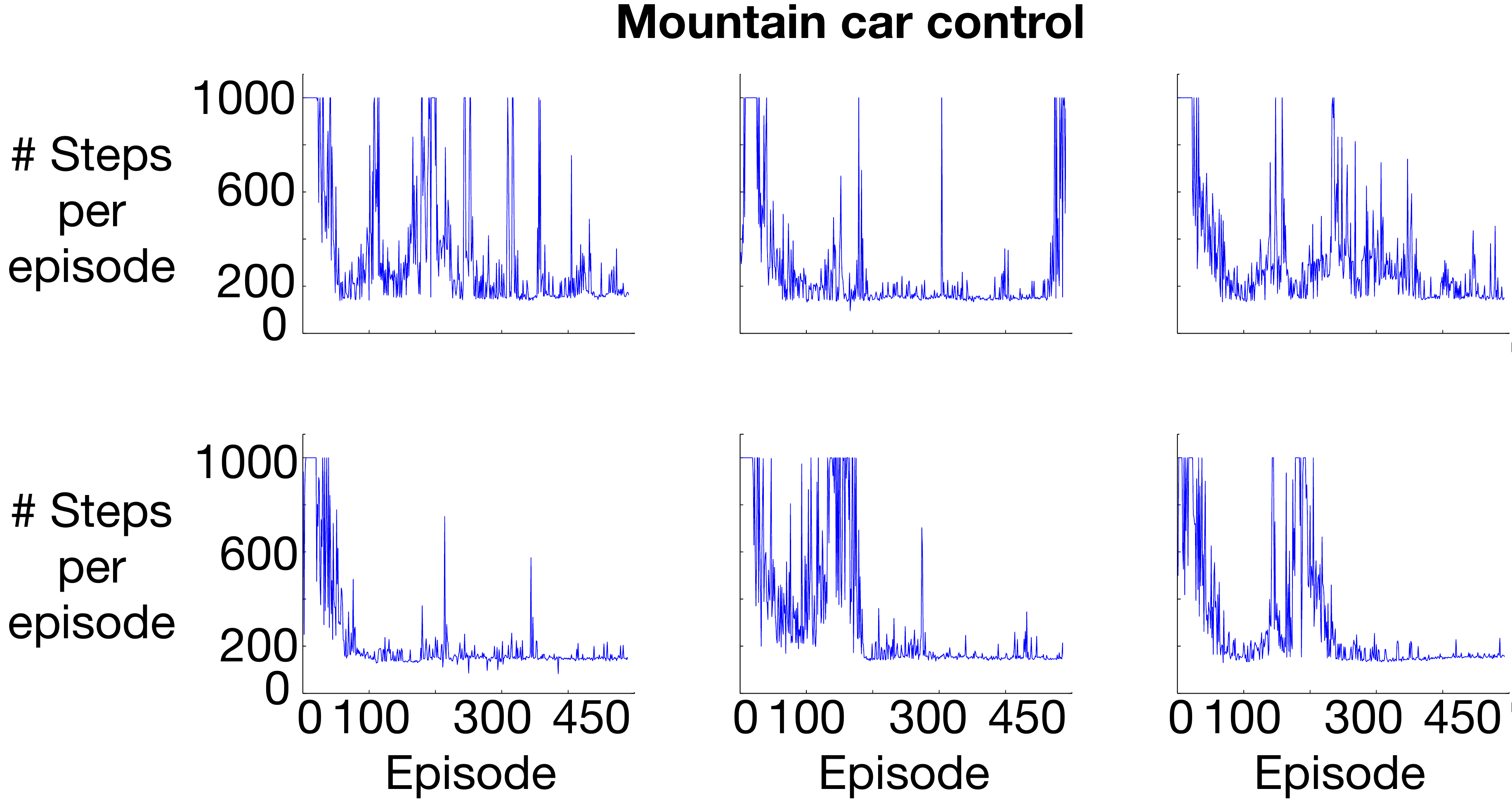} 
   \cprotect\caption{Sample learning curves of neural nets with $100$ RBF features, on the \verb+Mountain Car+ problem. Top row: troublesome runs; bottom row: better runs.}
   \label{fig:nnrbf_learning_int}
\end{figure} 

\begin{figure}[h]
  \centering
   \includegraphics[width=0.78\linewidth]{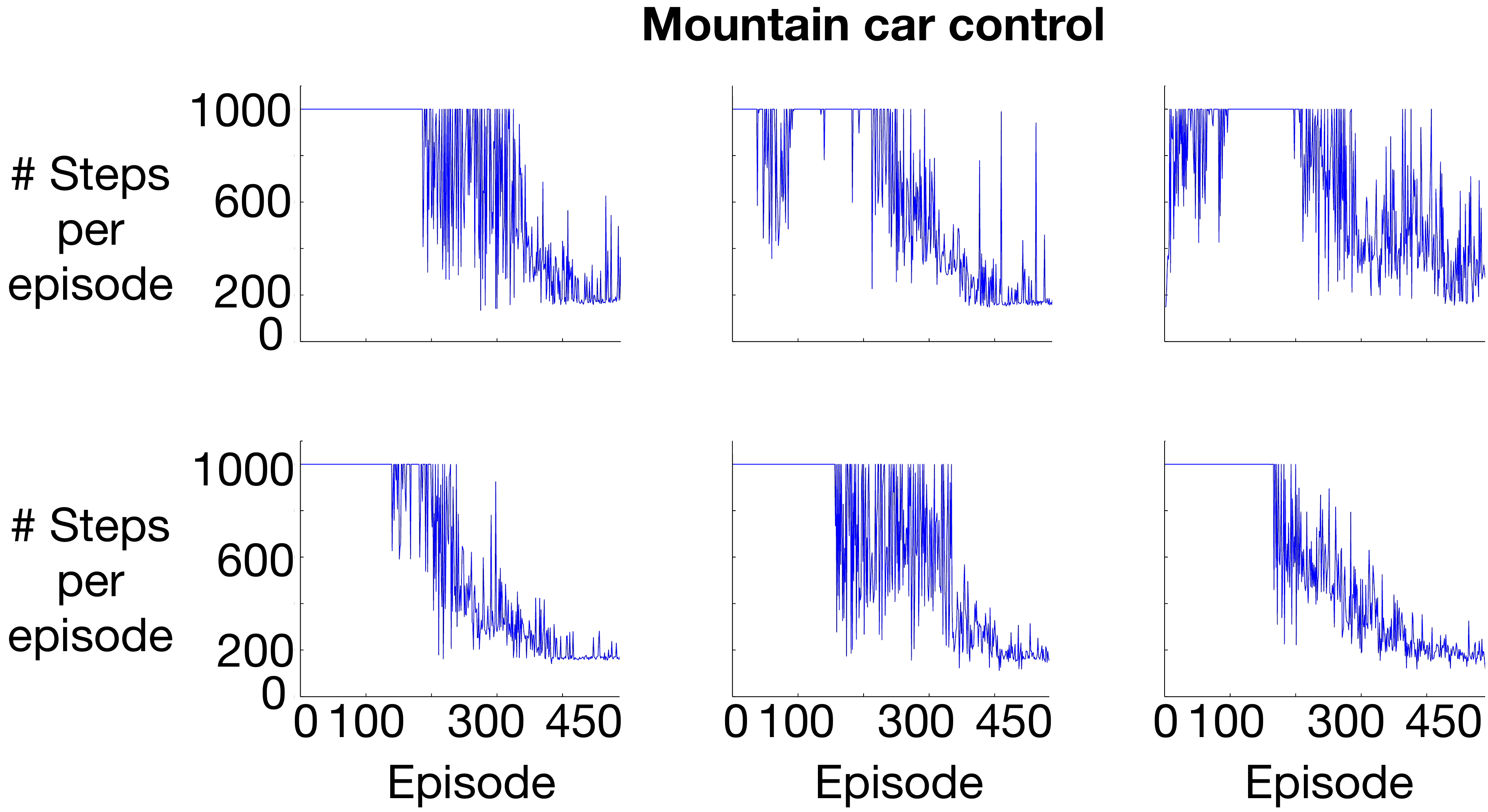} 
   \cprotect\caption{Sample learning curves of neural nets with $100$ sparsified RBF features, on the \verb+Mountain Car+ problem. Top row: troublesome runs; bottom row: better runs.}
   \label{fig:nnsrbf_learning_int}
\end{figure} 

\begin{figure*}[t]
  \centering
   \includegraphics[width=0.7\linewidth]{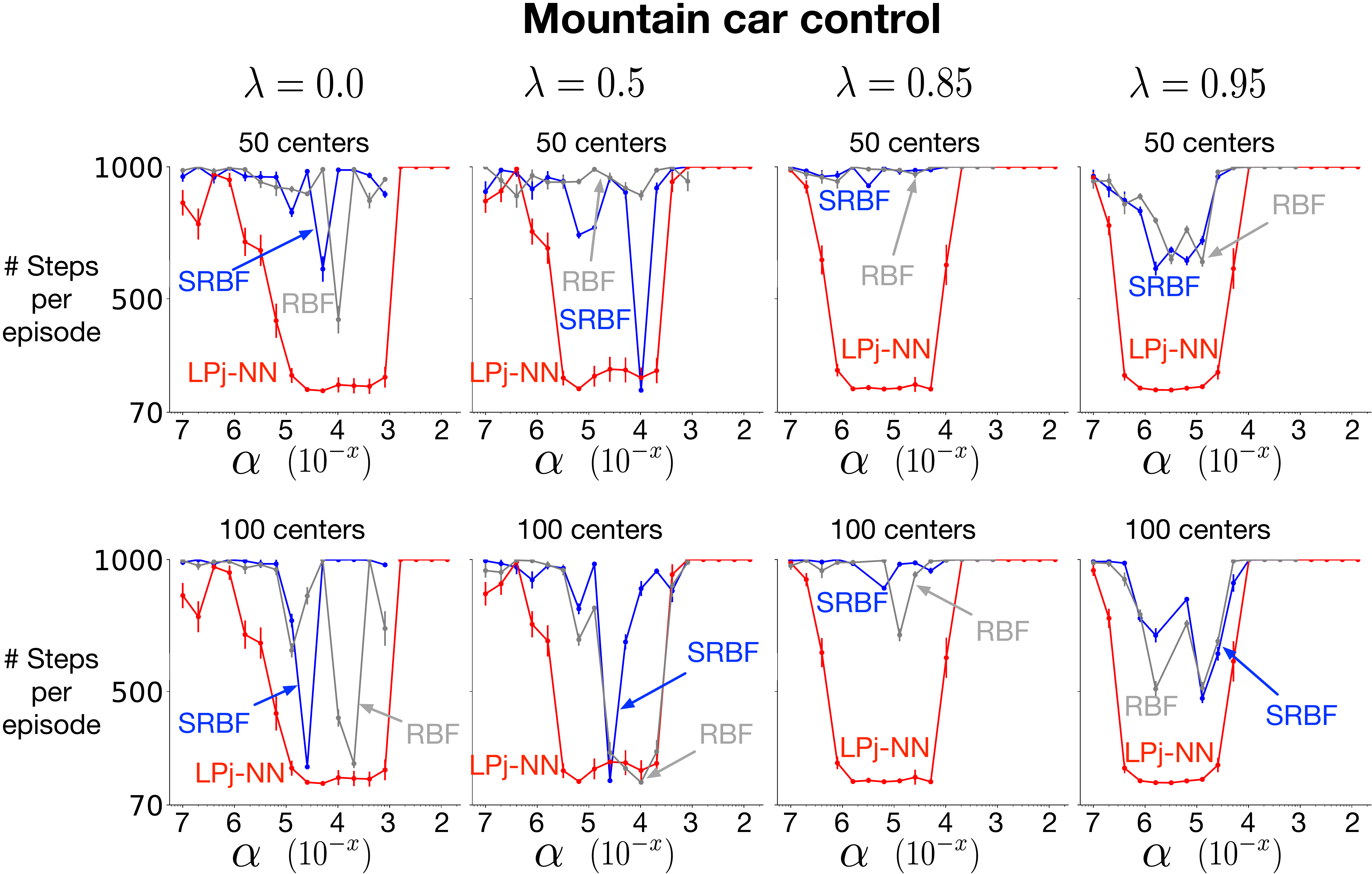} 
   \cprotect\caption{Parameter study for neural nets with $50$ and $100$ RBF or sparsified RBF (SRBF) features, on the \verb+Mountain Car+ problem. The performance of \Emj{} is plotted for comparison.}
   \label{fig:nn_rbf}
\end{figure*}

Figure \ref{fig:nn_rbf} shows the learning performance of neural nets for a range of step size parameters, when they operate on RBF or sparsified RBF (SRBF) features with width parameter $0.1$. (For the details about how we did this type of parameter study, see the experimental results section.) As can be seen, with RBF/SRBF features, neural nets suffered severe learning interferences, for both small and large $\lambda$ values.
 
The behavior of neural nets with RBF features shown above sharply contrasted with that of LPj-NN (which used only $3$ features). Note that in both cases, the input transformations involved are topological embeddings. The difference is that RBF features lack the other important aspect of EmECS: with radial basis functions, the inputs are not mapped to extreme points of a convex set. Instead they lie on a low-dimensional surface which can be so curvy that it becomes hard for individual neural nodes with ReLU gates, whose activation regions are half-spaces determined by hyperplanes, to separate a local region on this curvy surface from other parts of the surface (cf.\ Figure~\ref{subfig:rbf-featmap}).

Similarly, the behavior of neural nets with sparsified RBF features is in sharp contrast with that of TC-NN. Now in both cases, the input transformations create sparse features based on the geometrical structure of the state space. Again, the difference is that the sparsified RBF features do not lie among extreme points of a convex set, whereas the tile-coded features do and this, we believe, has helped reduce learning interferences and improved considerably the performance of the neural net.

Although it is hard to analyze the complex behavior of the entire network, to partially verify that what we just pointed out is a major reason why these neural nets with RBF/SRBF features failed, we plotted in Figure~\ref{fig:rbf-featmap} the response functions of several nodes, after training these neural nets on the \verb+Mountain Car+ control problem. 
The majority of the node response functions we found are global. Even the relatively local ones tend to have strong responses to multiple spots in the state space that are far apart from one another.
We plotted some of them in the two columns on the left side of the figure. 

These node response functions from neural nets with RBF/SRBF features can be contrasted with the node response functions of neural nets with tile coding or EmECS shown by Figure~\ref{fig:feature_map_TCNN} in the appendix on comparison of TC-Lin and TC-NN and Figure~\ref{fig:LPNNFeatureMaps} in the appendix on comparison between EmECS and coarse coding.

\begin{figure*}[thb] 
   \centering
   \subfloat[RBF-NN\label{subfig:rbf-featmap}]{%
       \includegraphics[width=0.395\textwidth]{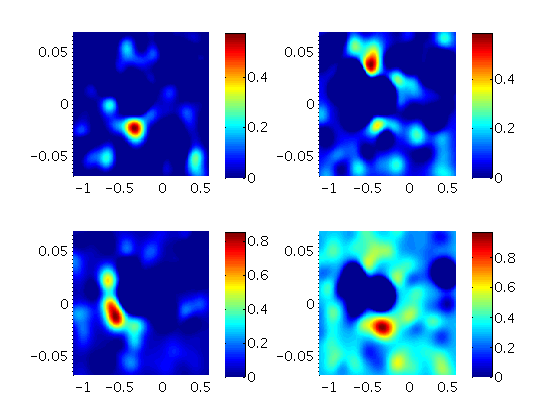}
       \includegraphics[width=0.395\textwidth]{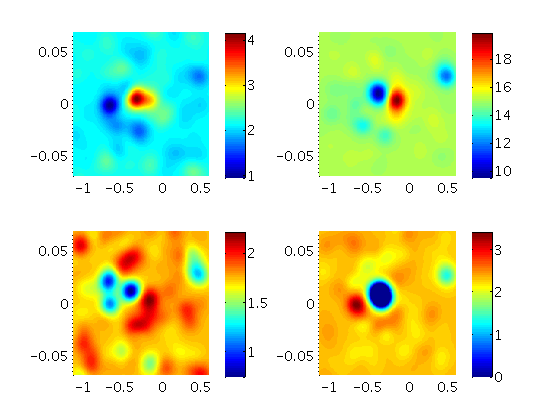}
     }
     \hfill
     \subfloat[SRBF-NN\label{subfig:srbf-featmap}]{%
       \includegraphics[width=0.395\textwidth]{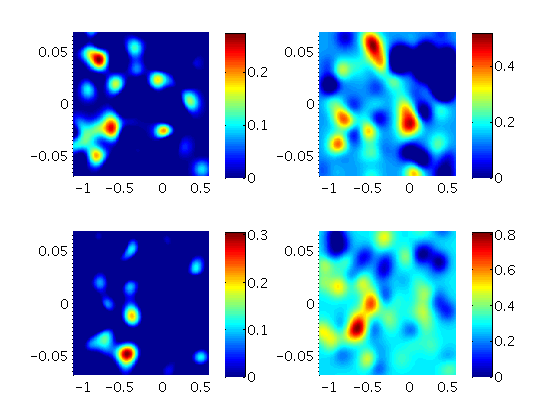}
       \includegraphics[width=0.395\textwidth]{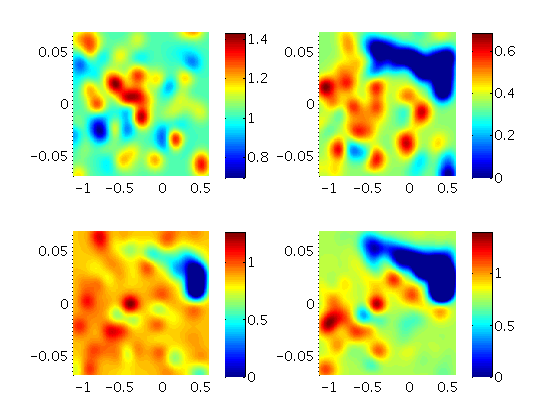}
     }
   \cprotect\caption{Response functions of neural nodes visualized on the original state space, for two neural nets operating on $100$ RBF or SRBF features, after training them on the task \verb+Mountain Car+. The horizontal (vertical) axis in the plots is position (velocity). The color scheme is such that if we normalize the responses to be between $0$ and $1$, dark red is $1$, green is $0.5$, and dark blue is $0$.} \label{fig:rbf-featmap}
 \end{figure*}

\clearpage
\section{Larger values of trace parameter $\lambda$} \label{app:largerLambda}

In this appendix we provide the experimental results for larger values of the trace parameter $\lambda$, for three different prediction and control tasks
that we considered in Section~\ref{sct:Experimental_results}. More specifically, we  provide the results for $\lambda=0.5$ and $\lambda=0.95$ and for four algorithms: tile coding jointly with neural networks (\TCjNN{}), lift-and-project jointly with neural networks (\Emj{}), tile coding with linear function approximation (\TCjLin{}), and neural networks with raw inputs (\NN{}).
The results are shown in Figure~\ref{fig:MCP_large_lambda}. We also performed experiments with $\lambda=0.85$, the results of which are not shown here because they were similar to $\lambda=0.95$.

As can be seen from these figures, although all methods got better in terms of speed and final performance, \NN{} was still slower than the other methods across all tasks. In \verb+Mountain Car+ prediction, \Emj{} and \TCjNN{} maintained their superior performance as larger values of $\lambda$ were used. In \verb+Mountain Car+ control, all methods had a similar final performance for each specific value of $\lambda$, except for \NN{} which could not solve the task. However, as $\lambda$ increased, each method achieved faster learning and better final performance, compared to when it used a smaller $\lambda$.
In the \verb+Acrobot+ task, \TCjNN{} outperformed other methods in terms of speed and final performance, when $\lambda = 0$ or $0.5$. When $\lambda=0.95$, the performance of all methods were close to each other. 

\begin{figure}[t]
      \centering
      \includegraphics[width=0.85\linewidth]{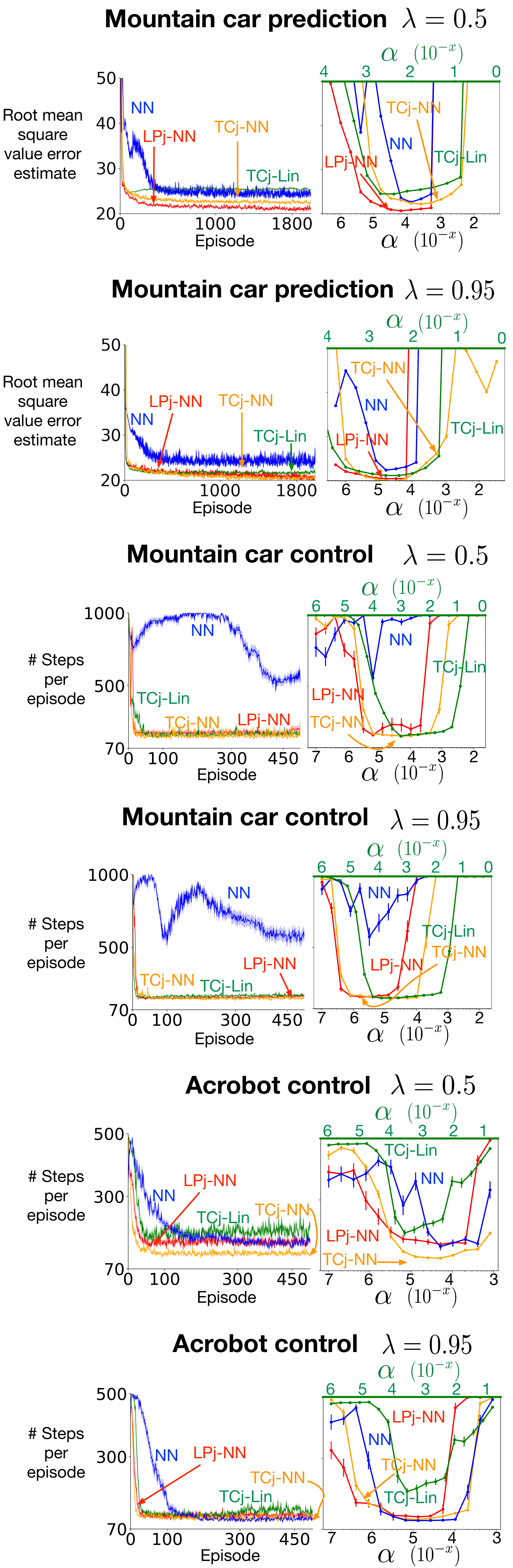}
      \caption{Learning curves and parameter studies for two different values of $\lambda$. }
      \label{fig:MCP_large_lambda}
\end{figure}

\clearpage
\clearpage
\section{Comparison of TC-Lin and TC-NN} \label{app:tile_coding}

Here, we continue our discussion from Section~\ref{sct:Tile_coding_plus_neural_networks} of the paper. We first show experimentally how TC-Lin and TC-NN are different. We then provide an intuitive explanation of why \TCsLin{} suffers from generalization restrictions whereas \TCsNN{} does not. After that we use neural node response function (similar to the ones in Appendix \ref{app:nn-int}) to explain why \TCsNN{} performs as well as \TCjNN{}.

We compare \TCsLin{}, \TCjLin{}, \TCsNN{}, and \TCjNN{} for all values of $\lambda$ on the \verb+Mountain Car+ prediction problem. (Results of \Ems{}, and \Emj{} are also shown). Our results in Figure~\ref{fig:joint_vs_separate_all_lambda} show that for all values of $\lambda$, \TCsLin{} is worse than \TCjLin{} in terms of asymptotic performance. \TCsLin{}'s range of step size for which it converges is smaller than \TCjLin{}. However, \TCsNN{} and \TCjNN{} (and also \Ems{} and \Emj{}) achieve the same performance for the same values of step sizes. These results show that TC-NN outperforms TC-Lin for all values of $\lambda$.

When we tile code different dimensions of the input space separately and combine them with linear function approximation, the resulting method (\TCsLin{}) has a rather critical limitation: it cannot take into account how features from different dimensions interact with each other. For example, in the continuous 2D space, it cannot express if a feature from the first dimension is good when the feature from the second dimension has a specific value. Whenever \TCsLin{} updates the weights corresponding to feature $x$ in the first dimension, it generalizes for \emph{$x$ and all} features of the second dimension, which might not be ideal. \TCjLin{} solves this problem by coding both dimensions at the same time and capturing the neighborhood information in both dimensions. Neural nets generalize differently and therefore do not encounter the aforementioned problem. A neural net with ReLU gates can choose to generalize for feature $x$ from the first dimension when feature $y$ from the second dimension has a specific value or, in the case of tile coding, when feature $y$ is absent (equal to zero) even if the input dimensions are tile coded separately. This is because in a fully connected network, all the features are gathered together at each node, and each node can respond to the features of its choice. The conclusion is that when the input space is tile coded separately, a linear function approximation method has a limitation that neural nets do not have.

Now let us focus on the neural node response functions for \TCsNN{} and \TCjNN{} and compare them to those of \NN{}. Figure~\ref{fig:feature_map_TCNN} shows some response functions for \TCjNN{} and \TCsNN{}. We followed the same procedure as in Appendix \ref{app:nn-int} to create these figures. The figures show that the response functions for \TCjNN{} and \TCsNN{} are similar, meaning that when the input is tile coded separately or jointly, the neural net still generalizes between states in a similar fashion.

\begin{figure}[H]
      \centering
      \includegraphics[width=0.8\linewidth]{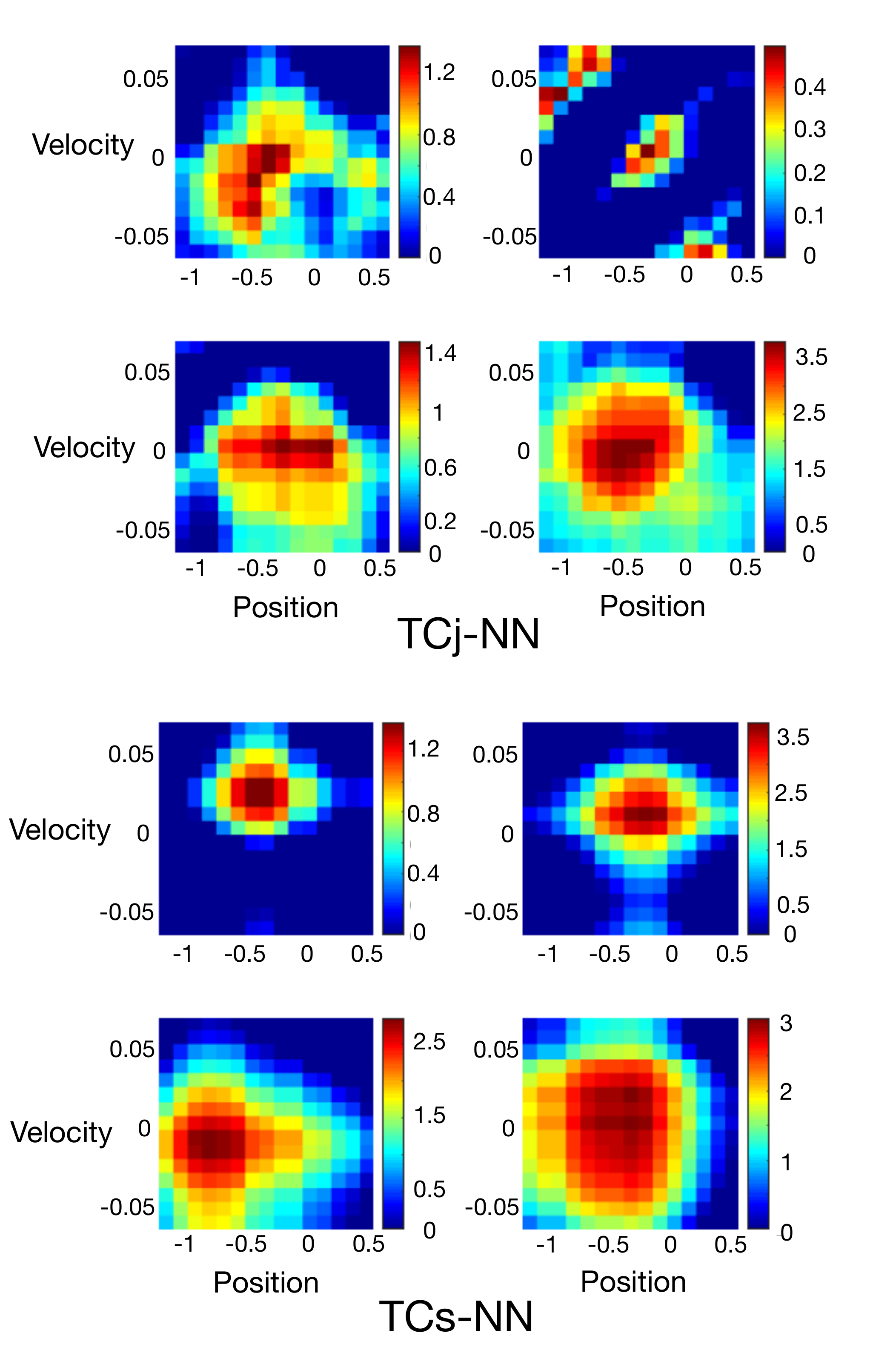}
      \cprotect\caption{Response functions for tile coding plus neural nets on the \verb+Mountain Car+ control problem.}
      \label{fig:feature_map_TCNN}
\end{figure}

\begin{figure}[H]
      \centering
      \includegraphics[width=0.8\linewidth]{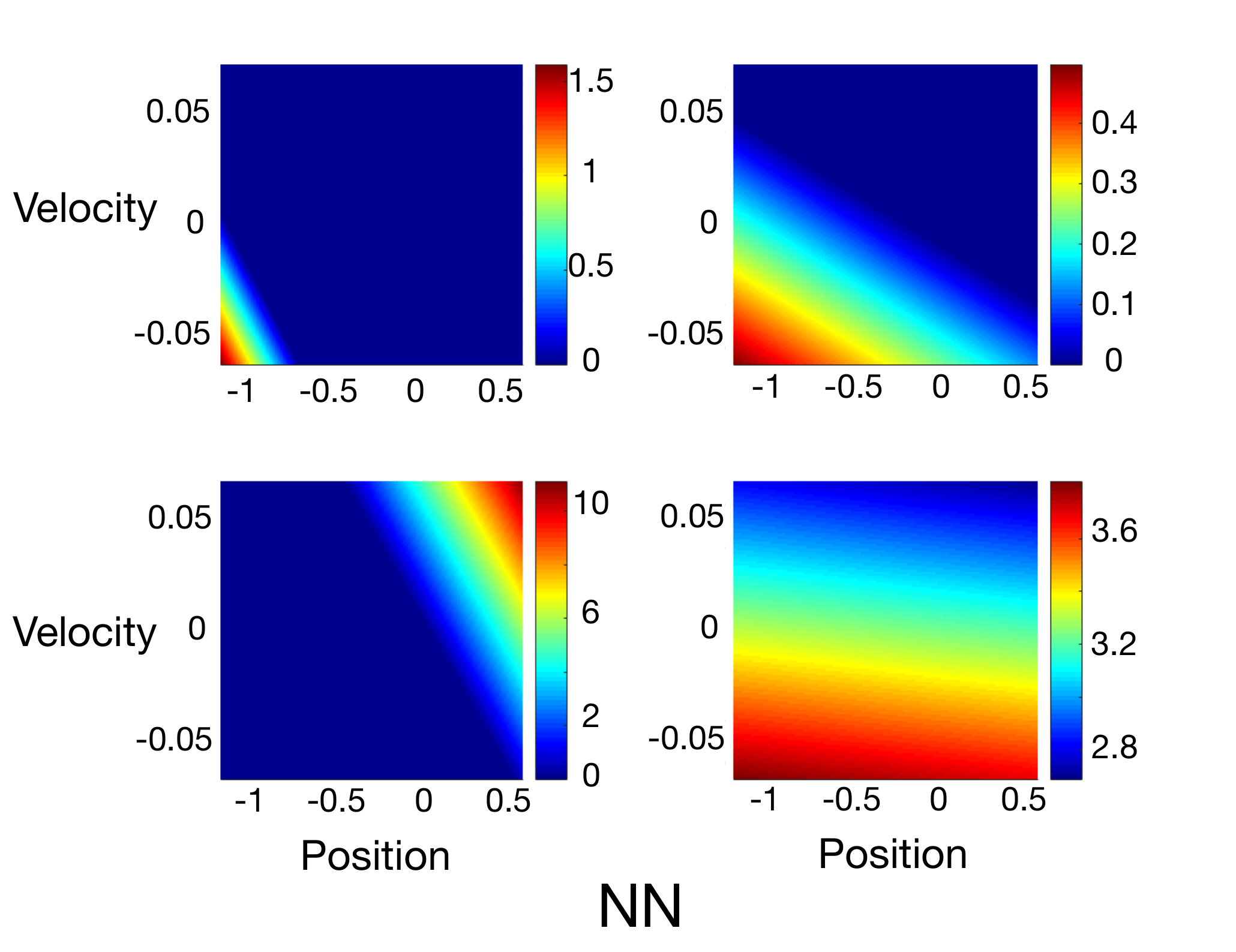}
      \cprotect\caption{Response functions for \NN{} with raw inputs for the \verb+Mountain Car+ control problem.}
      \label{fig:feature_map_NN}
\end{figure}

\begin{figure*}
      \centering
      \includegraphics[width=0.7\linewidth]{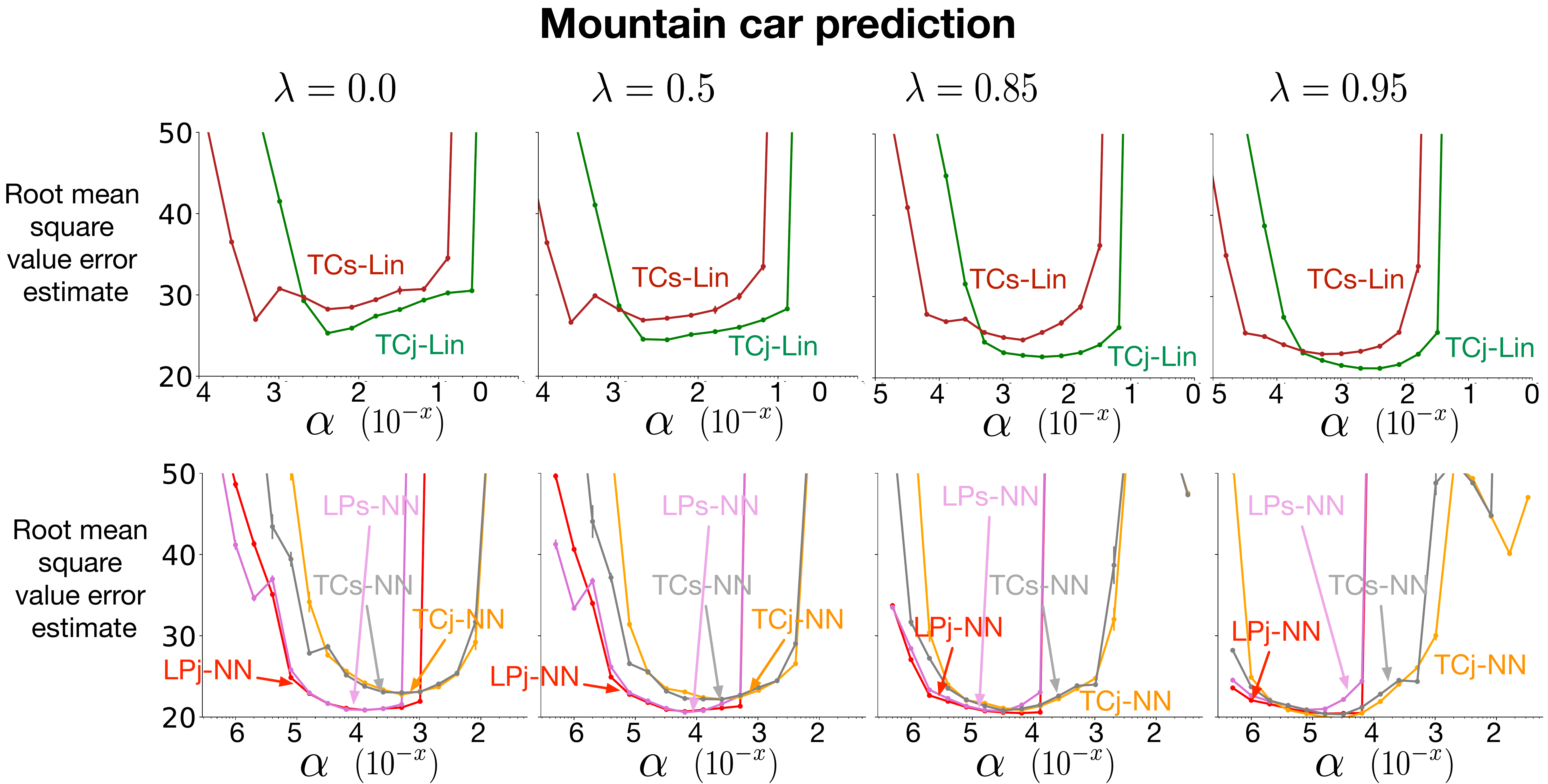}
      \caption{Parameter studies for different values of $\lambda$.}
      \label{fig:joint_vs_separate_all_lambda}
\end{figure*}

One important characteristic of \TCjNN{} and \TCsNN{} is that the nodes tend to focus their responses to a connected region. The reason is that when the inputs are tile coded, they are mapped to vertices of a hypercube in a higher dimension. Although the state space does not preserve its shape in this transformation, the neighborhoods that are close in the original space, tend to have a small Hamming distance between their binary representations and therefore they are still close to each other on the hypercube. This means that the nodes respond to neighborhoods that are close in the original space. However, there can still exist nodes that respond to different regions of the space (see the top right figure of \TCjNN{} in Figure \ref{fig:feature_map_TCNN}).

When response functions from \TCjNN{} and \TCsNN{} are compared to the ones for \NN{} in Figure~\ref{fig:feature_map_NN}, one can observe that the response of a single node in NN is linear within its activation region while \TCjNN{} or \TCsNN{} approximate a more complex function that is not linear with respect to the original input space in their activation regions.

\clearpage
\section{Comparison between EmECS and coarse coding} \label{app:emex-cc}

In Section~\ref{sct:EmECSMethod} we briefly discussed some similarities between EmECS and coarse coding and how the choice of the embedding in EmECS can influence the shapes of the activation regions of nodes in the neural nets when they operate on transformed inputs. This appendix is a longer version of that discussion, with more details and illustrations.

In terms of local generalization, EmECS is similar in spirit to coarse coding, despite their being seemingly unrelated. As explained in Section~\ref{sct:Tile_coding_plus_neural_networks} and illustrated in Figure~\ref{fig:cc}, in coarse coding we cover the state space with overlapping sets (circles, in this case), referred to as receptive fields (of the corresponding features). Observations at state $s$ will activate the three shaded receptive fields that contain $s$. The union of these fields delineates the region in which generalization occurs, and this region is connected and composed of ``nearby'' states. 

When we apply EmECS, what resemble the generalization regions in coarse coding are the activation regions of each neural node, viewed in the original input space. Because of the embedding property, these activation regions are also connected (like in coarse coding), if the hyperplanes associated with the nodes have their normal vectors point in the right directions in the transformed input space, as we mentioned in Section~\ref{sct:EmECSMethod}. For instance, with the lift-and-project map in Example~\ref{ex-1}(a), it suffices that these normal vectors point ``upwards'' with respect to the extra $(n+1)$-th dimension, i.e., the $(n+1)$-th component of every normal vector is nonnegative. Similarly, if we use the separate embedding scheme described in Example~\ref{ex-1}(c) with lift-and-project maps, it suffices that those normal vector components associated with the extra dimensions are nonnegative.

If the physical state space is not $X$ but a connected subspace of $X$, then a connected activation region in $X$, when it is too large, can still contain disconnected sets of physical states. However, by the geometric property of extreme points discussed at the beginning of Section~\ref{sct:EmECSMethod} and by the embedding property, EmECS ensures that as the activation region (viewed in $X$) becomes sufficiently small, it will contain only a connected subset of physical states.

If a node is activated by an input and the network is updated with gradient-descent using that input and a small step size, the activation region as well as the response of the node within that region will be modified slightly.
Thus when carrying out generalization, both coarse coding and the neural nodes in EmECS respect the topological structure of the original input space, although coarse coding does that in a coarser way.

\begin{figure}[htb] 
   \centering
   \includegraphics[width=0.25\textwidth]{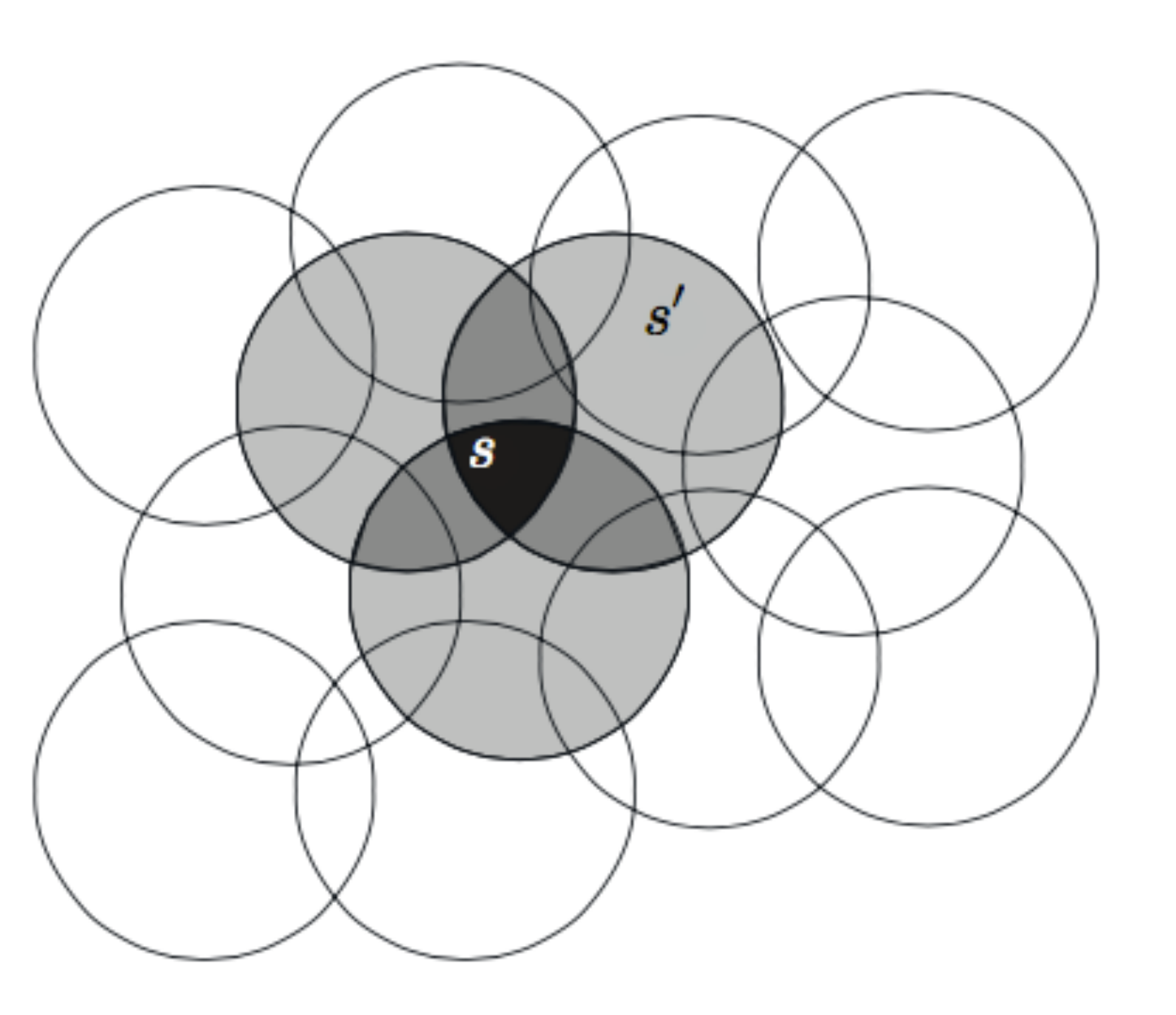} \qquad \qquad
   \raisebox{0.7cm}{\includegraphics[width=0.2\textwidth]{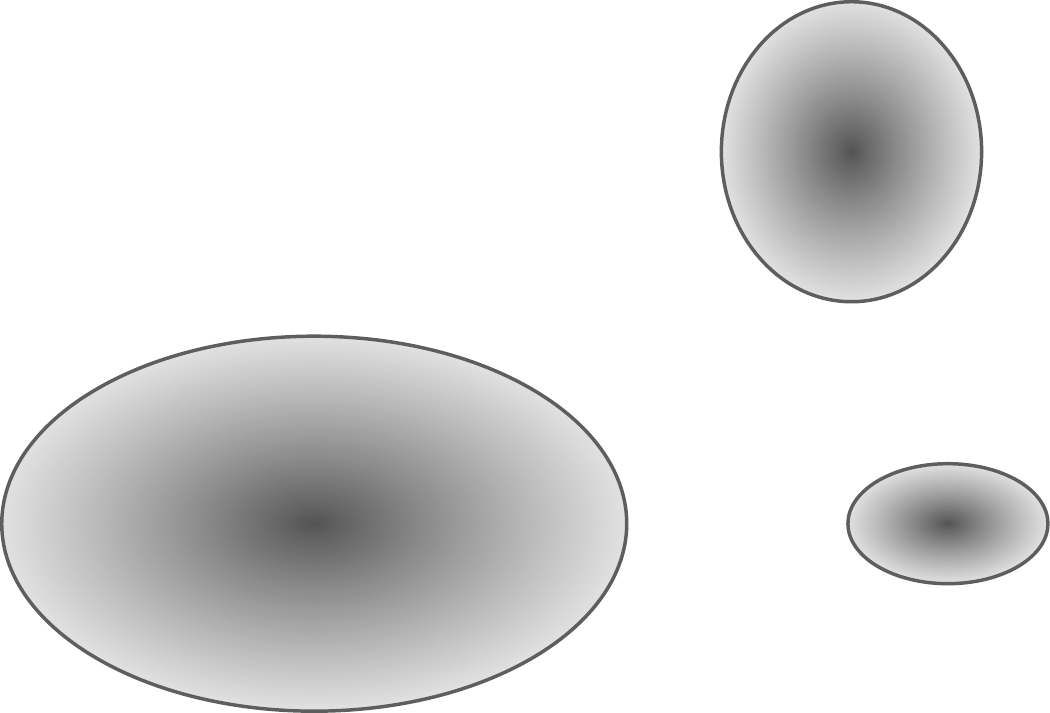}}
   \caption{Similarities between receptive fields in coarse coding and activation regions in EmECS. Top: receptive fields (in this case, circles) in coarse coding. Bottom: activation regions (ellipses, in this case) of three neural nodes in EmECS, viewed in the original input space. Each node responds with greater intensity (darker shading) to inputs lying deeper inside its activation region.}
   \label{fig:cc}
\end{figure}

In coarse coding, receptive fields can have different sizes and shapes. So are the activation regions in EmECS, as illustrated schematically in Figure~\ref{fig:cc}. However, the receptive fields in coarse coding are chosen in advance before learning takes place, whereas in EmECS, the activation regions change their locations and sizes dynamically during learning. Their shapes depend partly on the embedding, so by choosing the embedding, we can have influence over them, like in coarse coding. 

As a simple example, if we scale the coordinates of $X$ to create the space $X'$ and then embed $X'$ using the lift-and-project map in Example~\ref{ex-1}(a), then viewed in the original space $X$, the activation regions will be asymmetric, wider in some dimensions and narrower in others, and their shapes will also depend on where they are located.
As another example, the separate embedding scheme in Example~\ref{ex-1}(c) gives the network more freedom to produce activation regions that are asymmetric, whereas a joint embedding scheme like Example~\ref{ex-1}(a) can be used to enforce symmetric or specific asymmetric shapes.

Finally, to give a sense of what these activation regions actually look like when the neural nets have been trained on a task, we plotted in Figure \ref{fig:LPNNFeatureMaps} the response functions of a few neural nodes on the original input space, after the neural nets solved the \verb+Mountain Car+ control problem using Sarsa($0$). 
Two types of maps are used to apply EmECS in this experiment. The first one is the lift-and-project map given in Example~\ref{ex-1}(a); the corresponding algorithm is denoted LPj-NN.
The second one is the separate embedding scheme given in Example~\ref{ex-1}(c), where we embed the two dimensions of the state space separately, and we take the component maps $f_1, f_2$ to be lift-and-project maps given in Example~\ref{ex-1}(a). The corresponding algorithm is denoted LPs-NN.

The node response functions for LPj-NN and LPs-NN are plotted in Figure~\ref{fig:LPNNFeatureMaps}. The activation regions have different shapes under LPj-NN and LPs-NN. They are located across the state space, each focusing on some part of the space. However, instead of localized to small neighborhoods of particular states, they tend to be spread-out.  

\begin{figure*}[ht]
      \centering
      \includegraphics[width=0.7\linewidth]{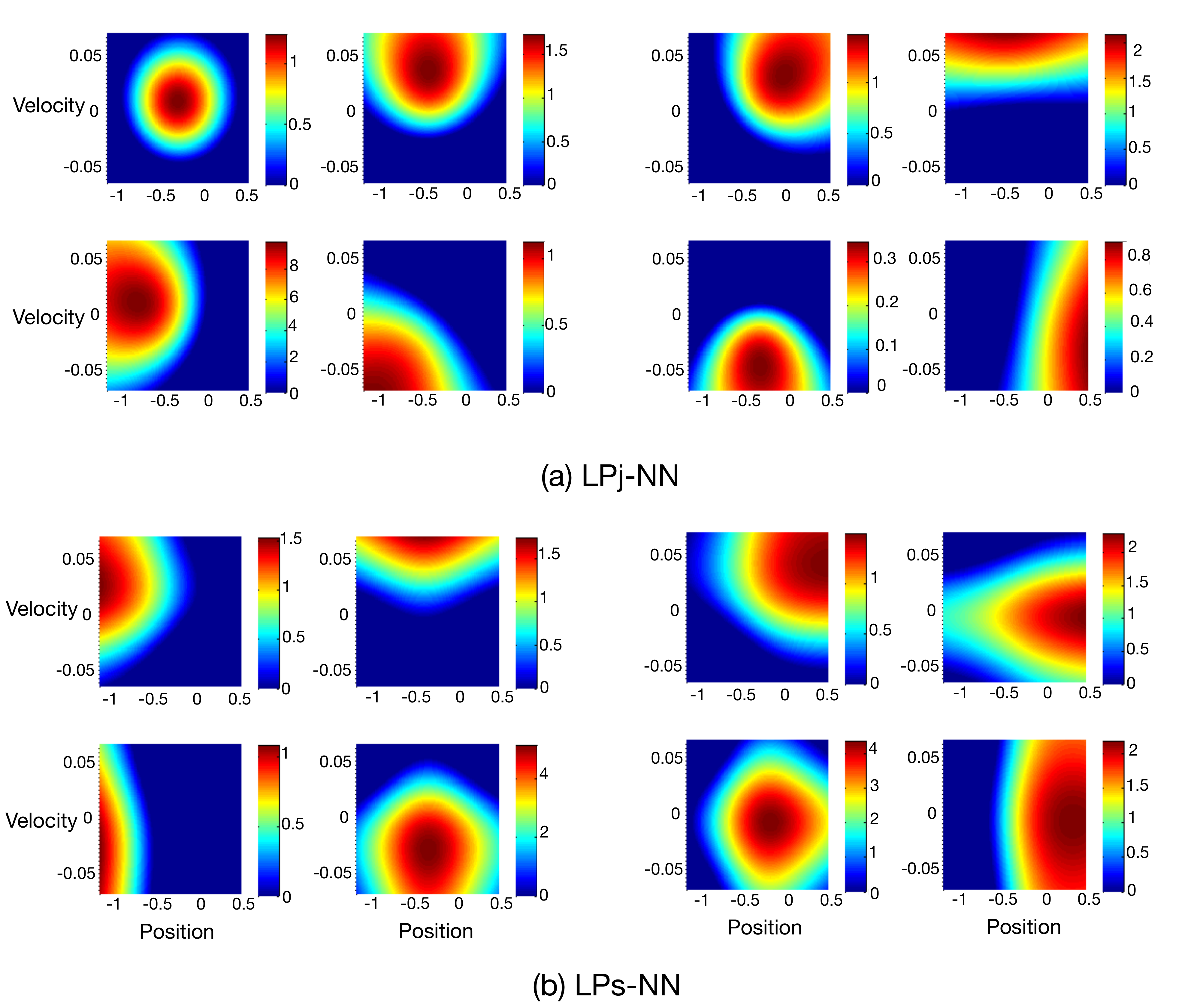}
      \cprotect\caption{Response functions of neural nodes visualized on the original state space, for two trained neural nets operating on inputs transformed by EmECS with joint (LPj) and separate (LPs) embedding schemes. The task is \verb+Mountain Car+; the horizontal (vertical) axis in the plots is position (velocity). The color scheme is such that if dark red represents $1$, then green is $0.5$ and dark blue is $0$.}
      \label{fig:LPNNFeatureMaps}
\end{figure*}

As to why these neural nets prefer large activation regions for each node, one explanation could be that each node was initialized with a random global activation region and it was hard for the nodes to coordinate with each other to shrink their activation regions to small neighborhoods of specific states. Another possible reason is that having many small activation regions makes it harder to approximate well the function values at the boundaries of the activation regions, and it is actually easier for the neural nets to get good approximations by having a spread-out response from each node. Further investigations are needed to better understand this behavior of neural nets.

\end{document}